\documentclass[journal]{IEEEtran}
\usepackage{enumitem}
\usepackage{soul}
\usepackage[T1]{fontenc} 
\usepackage[utf8]{inputenc}
\usepackage[small]{caption}
\usepackage{booktabs} 
\usepackage{flushend} 

\usepackage{cmap}
\input{glyphtounicode}

\usepackage{graphicx} 
\usepackage[caption=false]{subfig}

\usepackage{mathptmx} 
\DeclareMathAlphabet{\mathcal}{OMS}{cmsy}{m}{n}
\usepackage{times}    
\usepackage[english]{babel}
\usepackage[protrusion=false]{microtype}
\usepackage{amssymb}
\usepackage{mathrsfs}
\usepackage{mathtools}
\usepackage{url}
\usepackage{pbox}

\usepackage{hyperref}
\usepackage{xspace}
\usepackage[linesnumbered,ruled,vlined,algonl]{algorithm2e}
\SetAlCapNameFnt{\footnotesize}
\SetAlCapFnt{\footnotesize}
\DontPrintSemicolon
\SetKwInOut{KwInput}{Input}
\SetKwInOut{KwOuput}{Output}
\SetKwInOut{KwParam}{Param}
\SetKwInOut{KwState}{State}
\newcommand{\TCP}[1]{\hfill\tcp*[h]{#1}}
\newcommand*{\eg}{\emph{e}.\emph{g}.,\@\xspace} 
\newcommand*{\ie}{\emph{i}.\emph{e}.,\@\xspace}
\newcommand*{\prop}[3]{\ensuremath{{(#1,#2){:}#3}}\xspace}
\newcommand*{\class}[2]{\ensuremath{{#1{:}#2}}\xspace}
\newcommand*{\deff}[2]{{\ensuremath{{\exists}#1}.\ensuremath{#2}}}
\newcommand*{\defff}[2]{{\ensuremath{#1}.\ensuremath{#2}}}
\newcommand*{\fuzzy}[2]{\ensuremath{{\langle#1,#2\rangle}}\xspace}
\newcommand*{\fuzzyprop}[4]{\fuzzy{\prop{#1}{#2}{#3}}{#4}}
\newcommand*{\fuzzyclass}[3]{\fuzzy{\class{#1}{#2}}{#3}}
\newcommand*{\fuzzydef}[3]{\fuzzy{\deff{#1}{#2}}{#3}}
\newcommand*{\reify}[2]{\onto{#1}\hspace*{.1em}\onto{#2}}
\newcommand*{\nTH}[1]{\ensuremath{#1\text{-th}}\xspace}%
\newcommand{\onto}[1]{\text{{\fontfamily{lmss}\selectfont {#1}}}}
\usepackage{stmaryrd}
\usepackage{trimclip}
\makeatletter
\DeclareRobustCommand{\shortto}{\mathrel{\mathpalette\short@to\relax}}
\newcommand{\short@to}[2]{\mkern2mu\clipbox{{.5\width} 0 0 0}{$\m@th#1\vphantom{+}{\shortrightarrow}$}}
\makeatother
\usepackage{tikz}  
\usetikzlibrary{external}
\usepackage{ifthen}
\newboolean{shouldLoadPlots} 
\setboolean{shouldLoadPlots}{true}%
\newcommand{\rawepstextikz}[3]{
    \ifthenelse{\boolean{shouldLoadPlots}}{%
        \centering%
        \includegraphics[width=#1\linewidth]{./figure/#2.pdf}%
    }{%
        \begin{tikzpicture}%
            \node at (0,0){%
                #3%
                \def\svgwidth{#1\linewidth}%
                \input{./figure/#2.eps_tex}%
            };%
        \end{tikzpicture}%
    }%
}%
\newcommand{\epstextikz}[2]{
    \rawepstextikz{#1}{#2}{\centering\footnotesize}
}%
\usepackage{scalerel}%
\usetikzlibrary{svg.path}%
\definecolor{orcidlogocol}{HTML}{A6CE39}%
\tikzset{%
    orcidlogo/.pic={%
        \fill[orcidlogocol] svg{M256,128c0,70.7-57.3,128-128,128C57.3,256,0,198.7,0,128C0,57.3,57.3,0,128,0C198.7,0,256,57.3,256,128z};%
        \fill[white] svg{M86.3,186.2H70.9V79.1h15.4v48.4V186.2z}%
        svg{M108.9,79.1h41.6c39.6,0,57,28.3,57,53.6c0,27.5-21.5,53.6-56.8,53.6h-41.8V79.1z M124.3,172.4h24.5c34.9,0,42.9-26.5,42.9-39.7c0-21.5-13.7-39.7-43.7-39.7h-23.7V172.4z}%
        svg{M88.7,56.8c0,5.5-4.5,10.1-10.1,10.1c-5.6,0-10.1-4.6-10.1-10.1c0-5.6,4.5-10.1,10.1-10.1C84.2,46.7,88.7,51.3,88.7,56.8z};%
    }%
}%
\newcommand\orcidicon[1]{\href{https://orcid.org/#1}{%
    \ifthenelse{\boolean{shouldLoadPlots}}{%
        \centering
        \includegraphics[width=2.5mm]{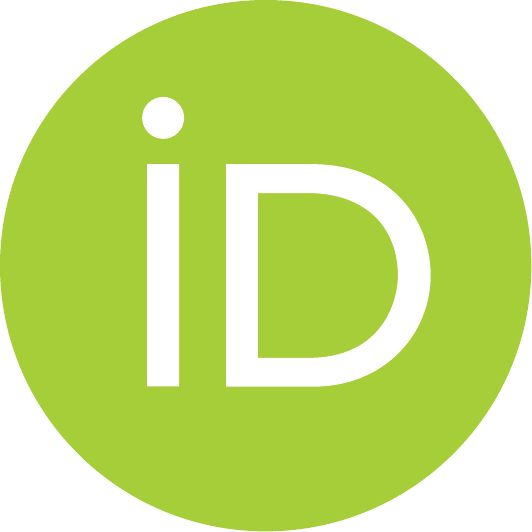}%
    }{%
        \centering
        \mbox{\scalerel*{\begin{tikzpicture}[yscale=-1,transform shape]\pic{orcidlogo};\end{tikzpicture}}{|}}%
    }}%
}%

\begin{document}%
\title{\LARGE Learning Symbolic Task Representation from a Human-Led Demonstration:\\A Memory to Store, Retrieve, Consolidate, and Forget Experiences}%
\ifthenelse{\boolean{shouldLoadPlots}}{%
    \author{Luca~Buoncompagni\textsuperscript{\orcidicon{0000-0001-8121-1616}} and Fulvio~Mastrogiovanni\textsuperscript{\orcidicon{0000-0001-5913-1898}}%
        \thanks{L. Buoncompagni and F. Mastrogiovanni
                are affiliated with the Department of Informatics, Bioengineering, Robotics and Systems Engineering, 
                University of Genoa, Via Opera Pia 13, 16145, Genoa, Italy.
                
                Corresponding author: \href{mailto:buon_luca@yahoo.com}{luca.buoncompagni@edu.unige.it}.
                
                Manuscript published the 16\textsuperscript{th} April 2024.}
    }%
}{
    \textbf{TODO} recompile while loading images from files to format authors, affiliations, and thanks (see the \emph{shouldLoadPlots} flag on the tex file).
}%
\newcommand{\keywords}{Structural Learning and Knowledge Capture, Fuzzy Description Logic, Learning Categories and Concepts, Logic-based learning algorithms, Learning by Demonstration, Human-Robot Interaction, Knowledge Representation and Cognitive Robotics, Cognitive Memory.}%
\markboth
         {L. Buoncompagni and F. Mastrogiovanni: Learning Symbolic Task Representation from a Human-Led Demonstration}{}
%
%
\makeatletter
    \hypersetup{hidelinks,
        pdfinfo={%
            Title={\@title},%
            Author={Luca Buoncompagni and Fulvio Mastrogiovanni},%
            Subject={A novel framework based on fuzzy ontology to structure robot's knowledge via Human demonstrations.},%
            Keywords={\keywords},
            pdfproducer={LaTeX},
            pdfcreator={pdfLaTeX}%
        }%
    }%
\makeatother%
\maketitle%
\begin{abstract} 
We present a symbolic learning framework inspired by cognitive-like memory functionalities (\ie storing, retrieving, consolidating and forgetting) to generate task representations to support high-level task planning and knowledge bootstrapping. 
We address a scenario involving a non-expert human, who performs a single task demonstration, and a robot, which online learns structured knowledge to re-execute the task based on experiences, \ie observations. 
We consider a one-shot learning process based on non-annotated data to store an intelligible representation of the task, which can be refined through interaction, \eg via verbal or visual communication. 
Our general-purpose framework relies on fuzzy Description Logic, which has been used to extend the previously developed Scene Identification and Tagging algorithm.
In this paper, we exploit such an algorithm to implement cognitive-like memory functionalities employing scores that rank memorised observations over time based on simple heuristics. 
Our main contribution is the formalisation of a framework that can be used to systematically investigate different heuristics for bootstrapping hierarchical knowledge representations based on robot observations. 
Through an illustrative assembly task scenario, the paper presents the performance of our framework to discuss its benefits and limitations.
\end{abstract} 
%
%
%
%
%

\thispagestyle{empty}

\section{Introduction}
\label{sec:introduction}

\noindent
\IEEEPARstart{T}{he} possibilities offered by human-robot collaboration are expected to significantly increase if robots can represent tasks from human-led demonstrations in an intelligible manner.
Since robots for in-home assistance or collaboration in factories should adapt to humans and their idiosyncrasies, they should be able to observe their actions and build a relevant knowledge representation.
Such representation must be structured and recalled when the robot needs to make decisions, but it should also be shared with humans, who could refine qualitative demonstrations by interaction.
In this scenario, the robot requires a knowledge representation that is transparent to the human and that should be generated from a few observations.
Furthermore, the method used for structuring knowledge should not be computed offline, and we should not assume that annotations are always available.

The paper presents a general-purpose framework to address the challenge of \emph{bootstrapping} a knowledge representation by interacting with humans and observing their actions~\cite{f2015Structurala}.
Our approach is inspired by a simplistic and functional model of memory, which is based on concepts broadly accepted in cognitive studies.
Memory processes and maintains information over time through three functionalities: \emph{encoding}, \emph{storing} and \emph{retrieving}~\cite{r2012review}.
In particular, sensory channels are used to encode information in some representation suitable for storage and retrieval within different types of memory, \eg short or long-term memory, as well as episodic or semantic memory~\cite{CogBotVernon}.
Due to limited resources, humans drive their attention to specific types of memory, and this is related to \emph{consolidation} (or \emph{rehearsal}) and \emph{forgetting}.

Rehearsal is a process that maintains information in the short and long-term memory~\cite{lockhart1990levels}, whereas consolidating involves episodic and semantic memory~\cite{consolEpisSema}.
Forgetting, which might be only temporally, is an essential and challenging behaviour for complex tasks since it allows focusing only on relevant aspects, and it might reduce ambiguities and inconsistencies.
Different theories have been proposed to model forgetting, \eg context-dependent strategies~\cite{s1988Environmental}, interference~\cite{b1957Interference} and trace decay~\cite{j2008Mind}.
In the first two, it is foreseen that some information may not be retrieved since stored items may be either corrupted or inconsistent.
According to the third one, retrieval is believed to be affected by time. 
This paper does not focus on an accurate model of the memory from a cognitive standpoint, but we argue that those functionalities can guide the design of effective robot memories for human-robot interaction scenarios. 

The paper presents a framework to bootstrap a hierarchical and symbolic knowledge representation based on the Description Logic (DL) formalism~\cite{baader2003description}.
The paper neither addresses perception issues, \ie we assume a perception module providing the robot with symbolic information, nor actuation, \ie the robot is a passive observer.
Instead, for the sake of argument, we developed a simple heuristic approach to learn a planning domain of an assembly task from a single demonstration, and we performed preliminary experiments.
Such a heuristic approach consolidates and forgets knowledge -- that is represented via symbols in an ontology -- by the means of scores-based ranking mechanisms. 
Symbols are acquired through the previously developed Scene Identification and Tagging algorithm (SIT), which is the base component of our framework and provides functions to encode, learn, structure and classify knowledge.
This paper exploits SIT to design a framework that implements storing, retrieving, consolidating and forgetting functionalities. 
Although our heuristics need further validation, our contribution is a formal framework that enables systematic investigations of different consolidating and forgetting strategies in support of storing and retrieving abilities. 

Our framework addresses scenarios where the knowledge that the robot bootstraps needs to be intelligible to a human supervisor, which might modify it through qualitative interaction and demonstrations.
Moreover, we consider scenarios where the robot acquires limited data that should be used to deduce a structured representation, \eg a plan concerning an assembly task. 
The SIT algorithm can bootstrap hierarchical representations, but it would also include irrelevant steps of the task plan since it tends to over-fit the demonstration.
Therefore, we extend SIT to consolidate the more relevant steps of a demonstration and forget the irrelevant ones. 

\section{Background}
\label{sec:relatedWork}

\subsection{Related Work}
In the literature, there are many data-driven approaches for learning probabilistic models of robot behaviour.
Typically, those methods do not provide structured and intelligible models, and they require many data samples to be processed in an offline training phase with the help of annotations.
Imitation learning has been successfully used to map raw data into primitive motions at a trajectory or symbolic level~\cite{a2008Robot}, using Reinforcement Learning~\cite{gu2017deep}, Deep Learning~\cite{t2018Deep} and Generative Adversarial Neural Networks~\cite{y2018Imitation}.
While those approaches could be viable within the Programming by Demonstration paradigm~\cite{a2008Robot}, they require a high amount of demonstrations, which are time-consuming for humans.
To speed up the training process, Meta-Learning has been used to transfer knowledge across domains, \eg using one-shot Imitation Learning based on Recurrent Neural Networks to model block-stacking tasks~\cite{y2017Oneshot}, and using a Convolutional Neural Network to learn actions from object locations~\cite{c2017OneShot}.
Probabilistic learning based on Hidden Markov Models has been used to generate intelligible and structured knowledge representations, \eg for motion planning~\cite{theocharous2001learning}, where humans could also interact within the controlling loop~\cite{lam2014pomdp}.
However, Markovian approaches do not bootstrap knowledge since they require the design of a prior structure for each application.
In addition, graph embedding is a suitable technique since planning and contextualised knowledge are often based on structures, and it also involves learning methods that process unstructured data~\cite{goyal2018graph}.

Symbolic representations have been used for task planning and knowledge bootstrapping not only because they support hierarchical structures, but also because they produce intelligible knowledge.
Symbolic and sub-symbolic models have been designed using the Ontology Web Language (OWL-DL) standard~\cite{tenorth2013knowrob} and the related \emph{reasoners}.
Given a symbolic representation of a domain, the problem of knowledge bootstrapping is to \emph{ground} observations, learn skills~\cite{konidaris2018skills}, and structure them into a representation, \eg a task plan.
With a similar perspective, Inductive Logic Programming has been used to learn OWL structures based on positive and negative examples~\cite{lisi2013logic}, and a technique for learning Hierarchical Task Networks from annotated datasets has been proposed in~\cite{hogg2016learning}.
However, those methods are used in an offline fashion and require annotated data.
In contrast, an approach to generate symbolic goals for a task planner through learned predicates and refinements, which are given from a supervisor only during the learning phase, has been proposed in~\cite{kirk2016learning}. 

From an analysis of the literature, it seems that the challenges related to the knowledge bootstrapping scenario we target have not been solved so far. 
Especially with the requirements of showcasing
($i$)~online and 
($ii$)~intelligible model learning (to support human supervision and refinements through interaction), 
($iii$)~few-shots learning from 
($iv$)~non-annotated data (to decrease human effort), and
($v$)~structured models learning (since it would effectively represent tasks as a sequence of actions).
Our framework addresses these requirements based on SIT functionalities, which are introduced in the next sections.

\subsection{Contribution}
\label{sec:contribution}
We used OWL-DL to develop the general-purpose SIT algorithm~\cite{oldSIT}, which performs one-shot learning and classification of scenes that are symbolically represented in a graph of crisp logic implications encoded in an OWL ontology. 
Although SIT bootstraps intelligible graphs, it is neither robust to vague representation nor to noisy inputs, and it tends to over-fit the demonstration because it might learn each observed scene.
Unfortunately, SIT does not scale efficiently since DL reasoning is known to be a complex task~\cite{baader2003description}.
Therefore SIT should be used in small scenarios, or in scenarios that can be represented across multiple ontologies~\cite{MON}.

In a previous work~\cite{l2018DialogueBased}, we deploy SIT in an application where a person talks with the robot to supervise and refine the knowledge it bootstraps during the demonstration of a task.
In that application, we exploited the \emph{supervisor} to overcome robustness and over-fitting issues of SIT, but the effort demanded to the human was consistent.
In order to reduce such an effort, we briefly introduced the idea of tackling over-fitting issues through scene consolidation and forgetting mechanisms based on a simulated crisp representation of knowledge~\cite{crisp_mem_ws}. 
Instead, to tackle the lack of robustness of SIT, we preliminary presented a fuzzy implementation of SIT~\cite{fuzzySit_ws, fuzzySIT2024}, which is an extension of SIT that exploits the \emph{fuzzyDL} reasoner~\cite{f2016fuzzy} to represent a structure of scenes through fuzzy-logic.
With respect to our previous work, this paper focuses on a framework that exploits the fuzzy implementation of the SIT algorithm, while implementing a scene consolidation and forgetting policy to overcome both robustness and over-fitting issues in a single implementation.

To present our framework, the next section briefly introduces the SIT algorithm, and Section~\ref{sec:fuzzyDL} presents the fuzzy ontology formalism used to implement the fuzzy extension of SIT%
    \footnote{The implementation is SIT based on a fuzzy ontology is available at \url{https://github.com/buoncubi/fuzzy_sit}.}, 
which is detailed in Section~\ref{sec:sit} as far as the objectives of this paper are concerned.
Our main contribution is presented in Section~\ref{sec:mem_cap}, which describes the knowledge storing and retrieving functionalities, as well as the assumptions and limitations of the consolidation and forgetting strategies that our framework can accommodate based on scores.
Section~\ref{sec:case2} discusses the preliminary results obtained from a toy -- but not simulated -- experiment based on simple consolidation and forgetting heuristics, which lead to the discussion of the benefits and limitations of our framework in Section~\ref{sec:discussion}.

\section{Overview}
\label{sec:overview}
SIT processes statements representing the environment, \ie \emph{facts}, based on a general-purpose interface (presented in Section~\ref{sec:input}), which formalises prior knowledge required to represent an observed \emph{scene}.
This interface involves symbols semantically representing the \emph{elements} in the scene (\eg objects of a certain \emph{type}) and the relationships (or \emph{roles}) between pairs of elements (\eg spatial relations).
Given this information, SIT performs \emph{encoding}, \emph{learning}, \emph{structuring} and \emph{classifying} functions that are summarised in Table~\ref{tab:SITphases}, which also introduces the main symbols defined in this paper.

Section~\ref{sec:encoding} presents the encoding functionality, which manipulates input facts to create \emph{beliefs} in the ontology, and computes their \emph{cardinality} to represent the scene.
For instance, a scene observed at a certain time can be represented with a belief as ``two cups are in front of a glass'' (Figure~\ref{fig:sceneEx}).

As detailed in Section~\ref{sec:learning}, the learning function generates scene \emph{categories} in the ontology, which are defined through \emph{restrictions} based on the cardinalities of the perceived beliefs.
For instance, a category can restrict a belief to have ``at least 2 cups in front of a glass''.
In this paper, we extend SIT to associate each category with a time-dependent \emph{score}, which is used to consolidate or forget such a category. 

The structuring function (addressed in Section~\ref{sec:structuring}) uses DL-based reasoning to generate a graph of categories based on their restrictions.
In particular, SIT creates a \emph{memory graph} (\eg Figure~\ref{fig:exGraph}) having scene categories as nodes, and edges as their logic implications.
At the beginning of the interaction, we assume a Lockean \emph{tabula rasa} situation, whereby the memory graph contains only a root node representing the empty scene, \ie it is restricted to ``at least'' zero beliefs.
Then, during the observation of a task, scene categories would be added to the memory graph for bootstrapping a knowledge representation.
Generally, the structuring function consistently arranges the structure of the graph when its categories are added or removed and, in this paper, we also assume that it preserves the score assigned to each category.

Given a memory, \ie a graph of categories, we can encode the input facts of the current scene for classifying its beliefs with respect to the categories previously learned and structured in the memory.
The classification function (presented in Section~\ref{sec:classifying}) provides a sub-graph of the memory, \ie the \emph{classification graph}, which contains all the categories defined with restrictions that satisfy the cardinality of the beliefs of a perceived scene, \eg Figure~\ref{fig:exGraph}.
Each node of the classification graph is related to a \emph{similarity value}, which describes how many restrictions are satisfied and, in the fuzzy implementation, it is also associated with a \emph{classification degree}, which represents how well the scene beliefs satisfy category restrictions. 

Section~\ref{sec:mem_cap} presents a combination of the above SIT functionalities used to implement a cognitive-like memory.
Section~\ref{sec:store} and Algorithm~\ref{alg:storeRetrieve} describe the design of \emph{storing} or \emph{retrieving} functions, while Section~\ref{sec:consolidate} and Algorithm~\ref{alg:consolidateForget} presents a strategy for \emph{consolidating} and \emph{forgetting} categories.
Although our algorithms are illustrative and implement simple heuristics, they support our contribution, which focuses on their input and output interfaces based on SIT.
Nonetheless, with such heuristics, we learn an effective representation of a table assembling task, as shown in Figure~\ref{fig:tableMount}.

To enable a discussion about storing, retrieving, consolidating and forgetting functionalities, Section~\ref{sec:sit} formalises the SIT functionalities introduced in this Section within an OWL ontology processed by a fuzzy DL reasoner.
The semantic associated with OWL and the logic-based reasoning for a fuzzy representation is briefly introduced in the next Section, and additional background material can be found in~\cite{baader2003description} and~\cite{f2016fuzzy}.

\begin{table}[t]
\centering
\vspace{.7em}
\footnotesize 
\newcommand{\PBP}[1]{\vspace{.1em}\pbox{.6\columnwidth}{\vspace{.1em}#1}}
\newcommand{\PB}[1]{\PBP{#1}\\\midrule}
\newcommand{\PBB}[2]{\vspace{.1em}\pbox{.2\columnwidth}{\relax\ifvmode\centering\fi\vspace{.1em}#1\\$\mathcal{#2}$}}
\newcommand{\PBBB}[1]{\vspace{.1em}\pbox{.2\columnwidth}{\relax\ifvmode\centering\fi\vspace{.1em}#1}}
\begin{tabular}{@{}c@{~~}r@{$\,\shortto\,$}l@{~~}c@{~}l@{}}
\toprule
Function                  & In                                    & Out                & \multicolumn{1}{c}{Description} \\
\midrule
\PBB{Encoding}{E}         & $F$                                   & $\epsilon$    
                   & \PB{Given \emph{facts} $F$\footnotemark[2], $\mathcal{E}$ reifies\footnotemark[3]~\emph{beliefs} $\mathbf{r}_{zh}$ and computes their \emph{cardinality} $c_{zh}$ to represent the perceived \onto{Scene} $\epsilon$.}
\PBB{Learning}{L}         & \PBBB{$\epsilon$\\[.2em]$q^0$\\[.2em]$a$}              & \PBBB{$\Phi_\epsilon$\\[.2em]$q_\epsilon$}  
                   & \PB{Given $\epsilon$, an \emph{initial score} $q^0$, and a \emph{fuzziness} value $a$, $\mathcal{L}$ computes cardinality \emph{restrictions} ${^{a}\Omega}$ defining a new scene \onto{CATEGORY} $\Phi_\epsilon$ with a \emph{score} $q_\epsilon$.}
\PBB{Structuring}{S}      & \PBBB{$M_{t-1}$\\[.2em]$\Phi_\epsilon$\\[.2em]$q_\epsilon$}  & $M_t$              
                   & \PB{Given a \emph{memory graph} $M_{t-1}$ in the ontology, and a new category $\Phi_\epsilon$ with a score $q_\epsilon$, $\mathcal{S}$ adds $\Phi_\epsilon$ and $q_\epsilon$ as a node of a new graph $M_t$, which consistently updates $M_{t-1}$ through logic reasoning.}
\PBB{Classifying}{C}      & \PBBB{$M_{t}$\\[.2em]$\epsilon$}                       & $M^\star$        
                   & \PBP{Given $\epsilon$ and $M_t$, $\mathcal{C}$ exploits logic reasoning to return the \emph{classification graph} $M^\star\subseteq M_t$, which contains the nodes $\Phi_j$ having $\epsilon$ as a realisation in the ontology (\ie \class{\epsilon}{\Phi_j}). Each node of $M^\star$ is also associated with the fuzzy \emph{classification degree} $p^{\class{\epsilon}{\Phi_j}}$ and the \emph{similarity value} $d^{\Phi_j}_\epsilon$.}\\
\bottomrule
\end{tabular}
\caption{The glossary of symbols and functionalities provided by the SIT algorithm. The latter is based on a fuzzy ontology and extended with \emph{scores} to implement consolidating and forgetting capabilities.}
\label{tab:SITphases}
\end{table}
\footnotetext[2]{Facts are element $\gamma$ of \onto{TYPE} $\Gamma$ and related with a \emph{role} $\mathbf{r}$.}
\footnotetext[3]{The reification operator is defined in Section~\ref{sec:encoding}.}

\section{A Fuzzy Description Logic Primer}
\label{sec:fuzzyDL}  
A DL ontology is a collection of logic statements (\ie \emph{axioms}, represented as tuples) describing knowledge through symbols with varied semantics.
In this work, we rely on the fuzzyDL reasoner, which logically deduces knowledge consistency based on a fuzzy degree of truth $p\in[0,1]$ assigned to each axiom.
Axioms are made of symbols related to two distinct ontological structures, \ie the Terminological Box (TBox), containing a hierarchical definition of \emph{concepts} and \emph{roles}, and the Assertion Box (ABox), including a set of \emph{instances}.
Furthermore, concepts are defined 
($i$)~as logical \emph{implications} with respect to other concepts, and 
($ii$)~through roles relating their \emph{definitions} to existing concepts. 
Roles are defined with mutual \emph{implications}, whereas instances can be described 
($i$)~as being a specific \emph{realisation} of a concept, and 
($ii$)~with \emph{assertion} roles in the definition of other instances.

Throughout the paper, we denote \emph{sets} with an upper case Roman letter, their \emph{items} or \emph{numbers} with lower case Roman letters, and \emph{functions} with Italic symbols.
Also, Table~\ref{tab:DLglossary} summarises the notation we adopt for the DL formalism.
The first column shows generic symbols for DL entities and the related fonts that we use to present examples, while the second and third columns summarise DL axioms and operators having the following definitions.

Let \fuzzyprop{\alpha}{\beta}{\mathbf{r}}{\;p^{\prop{\alpha}{\beta}{\mathbf{r}}}} be a tuple, where $\alpha$ and $\beta$ are instances, $\mathbf{r}$ is a specific role, and $p$ denotes the vagueness of this relation to be true for $\alpha$ and $\beta$.
For instance, the axiom \fuzzyprop{\onto{Basin}}{\onto{Jar}}{\onto{bigger}}{0.9} \emph{asserts} that a \onto{Basin} is bigger than a \onto{Jar} with a fuzzy degree of $0.9$.

Let \fuzzyclass{\alpha}{\Lambda}{\;p^{\class{\alpha}{\Lambda}}} be a tuple, where the instance $\alpha$ is defined as a \emph{realisation} of the concept $\Lambda$ with a certain degree of truth $p$, \eg \fuzzyclass{\onto{Basin}}{\onto{CONTAINER}}{0.7} or \fuzzyclass{\onto{Jar}}{\onto{CONTAINER}}{1}.

Also, let $\Lambda=\fuzzydef{\mathbf{r}}{\Upsilon}{\;p^{\defff{\mathbf{r}}{\Upsilon}}}$ be a tuple that \emph{defines} a concept $\Lambda$ with a degree $p$. 
Then, each realisation of $\Lambda$ is an instance $\alpha$ related with some (\ie $\exists$) role $\mathbf{r}$ to an individual $\beta$ that is a realisation of the $\Upsilon$ concept.
For instance, ${\onto{CONTAINER}=\fuzzydef{\onto{shape}}{\onto{CONVEX}}{1}}$ states that the instance \onto{Basin} is a consistent realisation of \onto{CONTAINER} only if in the ABox exist, with some degrees,
($i$)~the assertion \prop{\onto{Basin}}{\onto{Parabolic}}{\onto{shape}} and
($ii$)~the realisation \class{\onto{Parabolic}}{\onto{CONVEX}}.

In the TBox, axioms can be related to each other through set operators. 
In particular, we recall 
the \emph{disjunction} $\sqcup$, based on the fuzzy \emph{t-conorm}, 
the \emph{conjunction} $\sqcap$, identified with the \emph{t-norm},
and the \emph{subsumption} $\sqsubseteq$.
Subsumption (or \emph{supersumption} $\sqsupseteq$)  can be considered as an \emph{implication} (\ie $\Lambda\Rightarrow\Upsilon$), and we can write $\langle\Lambda\sqsubseteq\Upsilon, p^{\Lambda\sqsubseteq\Upsilon}\rangle$ if all the $\alpha$ realisations of $\Lambda$ are also instances of $\Upsilon$  with a degree $p$, \eg ${\langle\onto{CONTAINER}\sqsubseteq\onto{OBJECT}, 1\rangle}$.
In DL, the subsumption operator is also defined among two roles $\mathbf{r}$ and $\mathbf{p}$ but, in this paper, we consider this operation only to collect different roles with a unique semantic.

By checking the consistency of all the axioms and operators in the Tbox and ABox, the reasoner can 
($i$)~classify instances as realisations of concepts and 
($ii$)~generate a graph where nodes are concepts and edges are logic implications (\ie subsumptions). 
Remarkably, the reasoner complies with the \emph{open-world} assumption~\cite{baader2003description}, which is appropriate for robot scenarios, but more complex to handle.

\begin{table}[t]
\vspace{.7em}
\centering
\footnotesize 
\begin{tabular}{@{}r@{~}c@{~~} @{~}r@{~}c@{~~} @{~}r@{~}c@{}}
\toprule
\multicolumn{2}{@{~}c@{~}}{Entity} & \multicolumn{2}{@{~}c@{~}}{Axiom} & \multicolumn{2}{@{~}c@{~}}{Operator}   \\
\midrule
\onto{Instance}    & $\alpha$        & Assertion      & \prop{\alpha}{\beta}{\mathbf{r}}      & Disjunction   & $\Lambda \sqcup \Delta$      \\
\onto{CONCEPT}     & $\Lambda$       & Realisation    & \class{\alpha}{\Lambda}               & Conjunction   & $\Lambda \sqcap \Delta$      \\
\onto{role}        & $\mathbf{r}$    & Definition     & $\Lambda=\deff{\mathbf{r}}{\Upsilon}$ & Subsumption   & $\Lambda \sqsubseteq \Delta,\,\mathbf{r}\sqsubseteq\mathbf{p}$ \\
\bottomrule
\end{tabular}
\caption{%
    The glossary of the DL-based symbols used to implement SIT. 
    The first column shows the notation we adopted to identify the different types of \emph{entities} in the ontology. 
    The second and third columns regard the used logic \emph{axioms} and \emph{operators} among entities.}
\label{tab:DLglossary}
\end{table}

\section{Scene Identification and Tagging}
\label{sec:sit}

For implementing the SIT functionalities summarised in Table~\ref{tab:SITphases}, we assume in the TBox the availability of two \emph{disjoint} concepts describing scenes \onto{CATEGORY} ($\Phi$, \ie the root of the memory graph), and the \onto{TYPE} ($\Gamma$) of elements in the scene.
The TBox also involves a \onto{role}~($\mathbf{r}$), while the ABox contains instances representing each \emph{element} of the environment (at least a pair $\gamma_x,\gamma_y$) and the current \onto{Scene}~($\epsilon$).
SIT represents observations that hold only for a discrete instance of time in the ABox, whereas it memorises a graph of scene categorisation over time in the TBox.

\begin{figure}[t]%
    \vspace*{-.7em}
    \centering%
    \footnotesize%
    \begin{minipage}{.32\linewidth}%
        \epstextikz{.95}{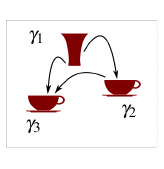}%
    \end{minipage}%
    \quad%
    \begin{tabular}{@{}l@{~~~~}l@{~}}%
         \fuzzyclass{\gamma_1}{\onto{GLASS}}{0.9}, & \fuzzyprop{\gamma_1}{\gamma_2}{\onto{front}}{1},    \\%
         \fuzzyclass{\gamma_2}{\onto{CUP}}{0.7},   & \fuzzyprop{\gamma_1}{\gamma_3}{\onto{front}}{0.6},  \\%
         \fuzzyclass{\gamma_3}{\onto{CUP}}{0.8},   & \fuzzyprop{\gamma_2}{\gamma_3}{\onto{front}}{0.2},  \\%
         \fuzzyclass{\gamma_3}{\onto{GLASS}}{0.1}, & \fuzzyprop{\gamma_2}{\gamma_1}{\onto{behind}}{1},   \\%
                                                   & \fuzzyprop{\gamma_3}{\gamma_1}{\onto{behind}}{0.6}, \\%
                                                   & \fuzzyprop{\gamma_3}{\gamma_2}{\onto{behind}}{0.2}. \\%
    \end{tabular}%
    \vspace{-1em}%
    \caption{%
        An example of input facts (on the right-hand side) required to process the spatial scene $\epsilon_1$ (on the left-hand side).
        Facts can be encoded by SIT to perform the functionalities shown in Table~\ref{tab:SITphases}.}%
    \label{fig:sceneEx}%
\end{figure}%

\begin{figure*}%
    \footnotesize%
    \centering%
    \begin{minipage}[c]{0.22\textwidth}%
        \vspace{1.5em}%
        \hspace{.3em}%
        \epstextikz{.9}{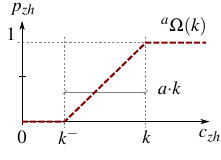}%
        \vspace{.3em}%
        \caption{%
            The fuzzy cardinality restriction $^{a}\Omega(k)$. 
            If the cardinality ${c_{zh} \geqslant k}$, the restriction is satisfied with fuzzy degree ${p_{zh} = 1}$. 
            If ${c_{zh} \leqslant k^-}$, it is not satisfied. 
            Otherwise, it is \emph{fuzzily} satisfied with ${p_{zh} \in (0,1)}$.
            We define ${k^{-} = k (1-a),\,a{\in}[0,1]}$.
        }%
        \label{fig:leftShoulder}%
    \end{minipage}%
    \hfill%
    \begin{minipage}[c]{0.75\textwidth}%
        \centering%
        \subfloat[$\epsilon_2$.]{%
            \raisebox{1em}{\epstextikz{.12}{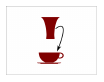}}%
            \label{fig:exScene2}%
        }%
        \subfloat[$\epsilon_3$.]{%
            \raisebox{1em}{\epstextikz{.12}{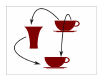}}%
            \label{fig:exScene3}%
        }%
        \subfloat[$\epsilon_4$.]{%
            \raisebox{1em}{\epstextikz{.12}{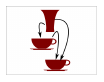}}%
            \label{fig:exScene4}%
        }%
        ~~~%
        \subfloat[The memory graph $M_t$.]{%
            \epstextikz{.29}{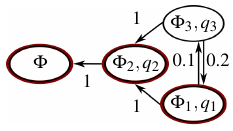}%
            \label{fig:exGraph}%
        }%
        ~~~%
        \subfloat[The classification of $\epsilon_4$.]{%
            \newcommand{\TN}[1]{\begin{tabular}[c]{@{}c@{}}#1\end{tabular}}
            \raisebox{3em}{
                \begin{tabular}{@{}c@{\hspace{1ex}}@{\hspace{1ex}}c@{}}
                    \toprule
                    \TN{Classification\\Degree}            & \TN{Similarity\\Value}               \\
                    \midrule
                    $p^{\class{\epsilon_4}{\Phi_1}}=0.8$   & $d^{\Phi_1}_{\epsilon_4}=0.9~~~$      \\
                    $p^{\class{\epsilon_4}{\Phi_2}}=1~~\,$ & $d^{\Phi_2}_{\epsilon_4}=0.4~~~$      \\
                    $p^{\class{\epsilon_4}{\Phi_3}}=0~~\,$ & $d^{\Phi_3}_{\epsilon_4}=\text{NaN}$  \\
                    \bottomrule
                \end{tabular}
            }
            \label{fig:exTable}%
        }%
        \caption{%
            Examples of scenes that are observed, learned and classified in the memory over time.
            When the $\epsilon_1$~(Figure~\ref{fig:sceneEx}), $\epsilon_2$~(\ref{fig:exScene2}) and $\epsilon_3$~(\ref{fig:exScene3}) scenes have been perceived, SIT learns and structures in the memory $M_t$ (\ref{fig:exGraph}) the categories $\Phi_1$, $\Phi_2$ and $\Phi_3$, respectively. 
            Each category is computed with $a = 0.5$, and it has a score (\ie $q_1$, $q_2$ and $q_3$).
            Then, when the $\epsilon_4$ scene~(\ref{fig:exScene4}) is perceived, it will be classified in a sub-graph of $M_t$, \ie the classification graph $M^\star$, which nodes are highlighted in~(\ref{fig:exGraph}). 
            Table~(\ref{fig:exTable}) shows the classification degree $p^{\class{\epsilon_4}{\Phi_j}}$ and similarity value $d^{\Phi_j}_{\epsilon_4}$ associates for each \nTH{j} node in $M_t$ except for $\Phi$, which is the root of the graph and it categorises an empty scene.
            Since $p^{\class{\epsilon_4}{\Phi_2}} = 0$, $\Phi_3$ is not a node of $M^\star$ and, therefore, $d^{\Phi_3}_{\epsilon_4}$ is undefined.
        }%
        \label{fig:SITex}%
    \end{minipage}
\end{figure*}

\subsection{The Input Interface}
\label{sec:input}
SIT requires a set $F$ of \emph{facts} ${\{f_1, \ldots, f_i, \ldots, f_n\}}$ describing the state of the robot workspace at a certain instant of time.
A generic fact $f_i$ is a set of DL axioms spanning in a combination of a DL entity subsuming 
($i$)~the \onto{role} ${\mathbf{r}\sqsupseteq\{\mathbf{r}_1, \ldots, \mathbf{r}_z, \ldots, \mathbf{r}_w\}}$,
which is applied between an element pair ($\gamma_x,\gamma_y$), where $\gamma_x$ and $\gamma_y$ are realisations of 
($ii$)~the \onto{TYPE} concept ${\Gamma\sqsupseteq\{\Gamma_1, \ldots, \Gamma_v\}}$.
Therefore, the sets of $w$ roles and $v$ concepts generally represent the prior knowledge used to define the input interface. 
Figure~\ref{fig:sceneEx} shows possible inputs where we consider an input interface with $\{\onto{front},\onto{behind}\} \sqsubseteq \mathbf{r}$ and  $\{\onto{CUP},\onto{GLASS}\}\sqsubseteq\Gamma$, \ie $w=v=2$. 

In general, given the concepts subsuming $\Gamma$, a role $\mathbf{r}_z$, and fuzzy membership degrees $p_{iz}, p_{is}, p_{ih} \in [0,1]$, we can formalise the input interface as a set of facts $F$ with items
\begin{align}
\label{eq:inputs}
f_i = \bigg\{\fuzzyprop{\gamma_x}{\gamma_y}{\mathbf{r_z}}{p_{iz}};\;
       &\Big\{\fuzzyclass{\gamma_x}{\Gamma_s}{\,p_{is}}, \forall \Gamma_s \sqsubseteq \Gamma\Big\};\notag\\
       &\Big\{\fuzzyclass{\gamma_y}{\Gamma_h}{\,p_{ih}}, \forall \Gamma_h \sqsubseteq \Gamma\Big\}\bigg\}.%
\end{align}
A fact $f_i$ describes 
($i$)~the relation $\mathbf{r}_z$ between the entities $\gamma_x$ and $\gamma_y$ with a degree $p_{iz}\triangleq p^{\prop{\gamma_x}{\gamma_y}{\mathbf{r}_z}}$ and
($ii$)~the type of the \nTH{x} and \nTH{y} entities as realisations of the \nTH{s} and \nTH{h} concepts subsuming $\Gamma$ with degrees ${p_{is}\triangleq p^{\class{\gamma_x}{\Gamma_s}}}$ and ${p_{ih}\triangleq p^{\class{\gamma_y}{\Gamma_h}}}$.
It is noteworthy that, due to the open-world assumption, an input fact $f_i$ might not specify $p_{is}$ and $p_{ih}$ for some types $\Gamma$ of the \nTH{x} and \nTH{y} elements.

\subsection{Encoding}
\label{sec:encoding}
SIT encodes \emph{beliefs} into an instance $\epsilon$ that represents the current \emph{scene} based on $F$.
This phase exploits the \emph{reification} operator ${\mathcal{R}(\mathbf{r}_z,\Gamma_h)\triangleq\mathbf{r}_{zh}}$ that defines a role through a concatenation of constituent symbols.
For instance, ${\mathcal{R}(\mathbf{r}_1,\Gamma_2)\equiv\reify{front}{GLASS}}$ represents the fact that a \onto{GLASS} $\class{\gamma_y}{\Gamma_2}$ has some element $\class{\gamma_x}{\Gamma}$ in \onto{front}, \ie $\mathbf{r}_1$.
In this paper, we simplify the reification $\mathcal{R}$ by disregarding the realisation of $\gamma_x$ in the $\Gamma_s$ concept.
Instead, we reduce $\gamma_x$ as an element of a generic \onto{TYPE} $\Gamma$.
 
This simplification of $\mathcal{R}$ reduces the computation complexity since the number of axioms in the ontology decreases, and this is significant because the reasoning task has an exponential complexity with the number of axioms~\cite{baader2003description}.
In addition, if we assume an input interface where each $\mathbf{r}_z\sqsubseteq\mathbf{r}$ have an \emph{inverse} role $\mathbf{r}_z^1\sqsubseteq\mathbf{r}$, we do not completely lose information about \class{\gamma_x}{\Gamma_s} since there will always be a $f_j\in F$ representing the inverse fact of $f_i$.
In other words, $f_j$ involves $\mathbf{r_z}^{-1}$ where the \nTH{s} and \nTH{h} concepts are swapped with respect to $f_i$. 
For this reason, in the example in Figure~\ref{fig:sceneEx}, we consider $\mathbf{r}_1=\mathbf{r}_2^{-1}$, \ie \prop{\gamma_x}{\gamma_y}{\onto{front}} is consistent only if the axiom \prop{\gamma_y}{\gamma_x}{\onto{behind}} exists with the same degree.
However, this simplification also reduces the intelligibility of the bootstrapped knowledge representation.

We define the encoding function $\mathcal{E}$ for a fact $f_i$ and for a set of facts $F$, respectively, as 
\begin{align}
\label{eq:perceiving} 
\mathcal{E}\left(f_i\right) &= \Big\{\fuzzyprop{\epsilon}{c_{izh}}{\mathbf{r}_{zh}}{1},\;   \forall z, h : f_i\Big\},\;\;\text{therefore},\notag\\
\mathcal{E}\left(F\right)   &= \Big\{\fuzzyprop{\epsilon}{c_{zh}}{\mathbf{r}_{zh}}{1},\;\;  \forall zh\in[1,w{{\cdot}}v]\Big\}.
\end{align}
Since in \eqref{eq:perceiving} we always define $p^{\prop{\epsilon}{c_{izh}}{\mathbf{r}_{zh}}}$ equal to $1$, a belief $\mathcal{E}\left(f_i\right)$ is a non-fuzzy (\ie crisp) role based on a combination of each \nTH{z} \onto{role} and \nTH{h} \onto{TYPE} occurring in $f_i$. 
The number of \nTH{zh} beliefs generated from an input interface is equal to all the combinations of the symbols subsuming $\mathbf{r}$ and $\Gamma$, \ie $w{{\cdot}}v$;
however, this amount can be lower due to the open-world assumption.
Since different facts can be mapped to the same \nTH{zh} belief, we represent with $c_{zh}\in\mathbb{R}^{+}$ the fuzzy \emph{cardinality} of the reified role $\mathbf{r}_{zh}$.

In the crisp formalisation of SIT, the cardinality is the number of equivalent roles applied to $\epsilon$, but this definition becomes ill-posed and challenging to define in a fuzzy domain~\cite{f2008Qualified}.
Among several definitions proposed for fuzzy counting, the $\sigma$-count is a popular approach that defines the cardinality with the sum of the fuzzy degrees of all the items in a set~\cite{j2014discussion}.
This approach treats cardinality as a real positive number, but it has a drawback since it computes the \emph{energy} of a set, which is considered to be a measure of the cardinality assuming that the higher the energy value, the more the number of elements such a set should have. 
This might not be true, and the $\sigma$-count cannot discriminate differences between a set having $10$ elements with fuzzy degree $0.1$, or one element with degree equal to $1$.

The cardinality of a belief $\mathbf{r}_{zh}$ is defined as the conjunction among all the realisation of $\gamma_x$ in $\Gamma$, which is in disjunction with the fact that $\gamma_y$ is a realisation of $\Gamma_h$ and it is paired with $\gamma_x$ through $\mathbf{r}_z$.
Thus, $c_{izh}$ is the fuzzy degree of the axiom
\begin{equation}
     \bigsqcup_{s}^{[1,v]} \Big\{ \fuzzyprop{\gamma_x}{\gamma_y}{\mathbf{r}_z}{p_{iz}} \sqcap \fuzzyclass{\gamma_x}{\Gamma_s}{p_{is}} \sqcap \fuzzyclass{\gamma_y}{\Gamma_h}{p_{ih}} \Big\}.
\end{equation}
This is computed with the DL Zadeh logic~\cite{f2016fuzzy} as
${c_{izh} = \max_{s}^{[1,v]}\{\min\{p_{iz},p_{is},p_{ih}\}\big\}}$, 
and with the $\sigma$-count approach we obtain
${c_{zh} = \sum_{i}^{[1,n]} c_{izh}}$.

In the example in Figure~\ref{fig:sceneEx}, the cardinality $c_{12}$ of the \reify{front}{GLASS} belief, and $c_{21}$ regarding \reify{behind}{CUP} are
\begin{align}
\small
\label{eq:ex1}
    c_{12} = & \max\big\{\min\{1,0.9,0.7\}\big\} + \max\big\{\min\{0.6,0.9,0.8\},        \notag\\
             & \min\{0.6,0.9,0.1\}\big\} = 1.3.                                             \\
\label{eq:ex2}
    c_{21} = & \max\big\{\min\{1,0.7,0.9\}\big\} + \max\big\{\min\{0.6,,0.8,0.9\}\big\} \notag\\
             & + \max\big\{\min\{0.2,0.8,0.7\}\big\} = 1.5.           
\end{align}
In other words, in $\epsilon_1$ ``$1.3$ objects are in front of the glasses'', and ``$1.5$ objects are behind the cups''.
It is worth noticing that the example in Figure~\ref{fig:sceneEx} involves the $\gamma_3$ element, which is a a realisation of both $\Gamma_1$ and $\Gamma_2$ types, \ie $\fuzzyclass{\gamma_3}{\Gamma_2}{0.1}$ due to perception error or vagueness.

\subsection{Learning}
\label{sec:learning}
Let an initial consolidating \emph{score} $q^0$ (described more in detail in Section~\ref{sec:store}) and the cardinality $c_{zh}$ of each belief encoded from $F$, the learning function $\mathcal{L}$ creates in the TBox a new \emph{category} $\Phi_\epsilon$ subsuming the root of the memory graph $\Phi$ (\ie the empty scene) by definition.

In particular, for each \nTH{zh} belief, $\Phi_\epsilon$ is defined through a fuzzy cardinality \emph{restriction} that is based on a left-shoulder membership function ${^{a}\Omega(k)}$, which has $k\coloneqq c_{zh}$, and it does not change over time.
Such a membership function is shown in Figure~\ref{fig:leftShoulder}, and it is based on the \emph{fuzziness} parameter ${a\in[0,1]}$, from which we derive ${k^{-} = k (1-a)}$.
If $a=0$, $\Omega$ would be a crisp restriction, while if $a=1$ the restriction would always be satisfied, even for small cardinality $c_{zh}$, but with a low fuzzy degree $p_{zh}$.
Let the cardinality $\bar{c}_{zh}$ of a \onto{Scene} $\bar{\epsilon}$ be encoded at a different instant of time, then if $\bar{c}_{zh}\geqslant k$ we consider the restriction to be fully satisfied, \ie $p_{zh}=1$.
Instead, if $\bar{c}_{zh}\leqslant k^-$ the restriction would not be satisfied since $p_{zh}=0$, while if $k^-<\bar{c}_{zh}<k$ we consider the cardinality restriction to be \emph{fuzzily} satisfied because $p_{zh}\in(0,1)$.

Based on this evaluation, the learning function creates a concept $\Phi_\epsilon$ (with an initial score) where \emph{at least} some \nTH{zh} beliefs must be observed to represent $\epsilon$.
We define the learning function as ${\mathcal{L}\left(\epsilon,q^0,a\right) = \langle  \Phi_\epsilon,\;q_\epsilon\coloneqq q^0\rangle}$ where
\begin{equation}
\label{eq:learning}
    \Phi_\epsilon=\fuzzy{\Phi}{1} \bigsqcap_{\substack{zh\\~}}^{\mathcal{E}(F)} \fuzzydef{\mathbf{r}_{zh}}{{^{a}\Omega(k)}}{1}.
\end{equation}
In accordance with Section~\ref{sec:fuzzyDL}, in \eqref{eq:learning} it is certain (\ie ${p^{\defff{\mathbf{r}_{zh}}{{^{a}\Omega(k)}}}=1}$) that, in order to assert a future scene $\bar{\epsilon}$ as a realisation of $\Phi_\epsilon$, its cardinality $\bar{c}_{zh}$ needs to satisfy all the \nTH{zh} restrictions ${^{a}\Omega(k)}$ of $\Phi_\epsilon$.
As seen above, such a \class{\bar{\epsilon}}{\Phi_\epsilon} realisation can have a fuzzy degree $p_{zh}$, which is based on $\bar{c}_{zh}$, $a$, and $k$; which is equivalent to the $c_{zh}$ cardinality related to $\epsilon$.
When \class{\bar{\epsilon}}{\Phi_\epsilon} occurs with a degree $p_{zh}>0$, we consider the scene $\bar{\epsilon}$ be classified in the $\Phi_\epsilon$ category, as better discussed in Section~\ref{sec:classifying}.
It is noteworthy that, by the definition of \eqref{eq:learning}, $\Phi_\epsilon$ always subsumes $\Phi$ with a crisp degree.

For instance, the category $\Phi_2$ learned from the scene $\epsilon_2$ in Figure~\ref{fig:exScene2} can be 
\begin{align}
    \Phi_2 = \langle\Phi,1\rangle & \sqcap \fuzzydef{\reify{front}{GLASS}}{{^{a}\Omega(0.9)}}{1} \notag\\
                                  & \sqcap\fuzzydef{\reify{behind}{CUP}}{{^{a}\Omega(1)}}{1}.
\end{align}
In other words, a future \onto{Scene} will be classified into the $\Phi_2$ category if it has ``at least 0.9 objects are in front of some glasses'' and ``at least 1 object is behind some cups''.

\subsection{Structuring}
\label{sec:structuring}
After each learning phase, SIT relies on the structuring function $\mathcal{S}$ to add a new category $\Phi_\epsilon$ and its score $q_\epsilon$ into the \emph{memory graph} $M_{t-1}$, which gets updated consistently with respect to the previously learned categories in a new memory graph $M_t$.
The new scene category is added as a node in the graph, and this operation is delegated to the fuzzyDL reasoner, which can also update the memory when a category is removed from the ontology, \ie forgotten.
In particular, the reasoner generates a graph $M_t$ with weighted edges based on the fuzzy degree of subsumption axioms involving each pair of categories in the memory, \ie $p^{\Phi_i\sqsubseteq\Phi_j}\;\forall\,\Phi_i,\Phi_j{\sqsubseteq}\Phi$, and we do not show the edge if $p^{\Phi_i\sqsubseteq\Phi_j}{=}0$.

Thus, to compute $\mathcal{S}$, it is sufficient to introduce in the TBox the newly created $\Phi_\epsilon$ concept and invoke fuzzyDL to perform instance checking, \ie it reasons on the comparison of the restrictions ${^{a}\Omega(k_i)}$ and ${^{a}\Omega(k_j)}$ for each \nTH{zh} belief of $\Phi_i$ and $\Phi_j$, respectively.
Let a fixed fuzziness $a = 0.5$ used to learn categories, such a reasoning process provides a degree 
\\[.3em]
\begin{tabular}{@{}r@{~~~}p{.9\linewidth}@{}}
    ($i$)   & $p^{\Phi_i\sqsubseteq\Phi_j}_{zh} = 1$, if $k_i^- \geqslant k_j^-$, 
              \eg ``at least $k_i = 1.4$'' always implies ``at least~$k_j = 0.8$''; or\\
    ($ii$)  & $p^{\Phi_i\sqsubseteq\Phi_j}_{zh} = 0$, if $k_i \leqslant k_j^-$,
              \eg ``at least $k_i = 0.6$'' does not imply ``at least $k_j = 1.8$''; or\\
    ($iii$) & $p^{\Phi_i\sqsubseteq\Phi_j}_{zh} \in (0,1)$ linearly to $k^-_i < k^-_j < k_i$,
              \eg ``at least $k_i = 0.75$'' implies ``at least $k_j = 1$'' with $p^{\Phi_i\sqsubseteq\Phi_j}_{zh} = 0.5$.
    \notag
\end{tabular}
\\[.3em]
Since \eqref{eq:learning} is based on conjunctions, each edge of $M_t$ has a weight $p^{\Phi_i\sqsubseteq\Phi_j}$ computed by the reasoner as the minimum $p^{\Phi_i\sqsubseteq\Phi_j}_{zh}$ degree among all the \nTH{zh} restrictions of $\Phi_j$.

As an  example, Figure~\ref{fig:exGraph} shows how the categories learned from $\epsilon_1$ (in Figure~\ref{fig:sceneEx}), $\epsilon_2$ and $\epsilon_3$ (in Figure~\ref{fig:exScene2} and \ref{fig:exScene3}) are structured in the memory graph as $\Phi_1$, $\Phi_2$ and $\Phi_3$, respectively.
In particular, $\Phi_2$ is related to the scene that restricts the smaller number of beliefs, thus it is a child of the root $\Phi$.
Since $\epsilon_2$ is a sub-scene of $\epsilon_1$ and $\epsilon_3$, $\Phi_1$ and $\Phi_3$ are children of $\Phi_2$, \ie $\Phi_1\sqsubseteq\Phi_2$ and $\Phi_3\sqsubseteq\Phi_2$ with degree 1.
Hence, if the restrictions of $\Phi_1$ and $\Phi_3$ are satisfied, then the restrictions of $\Phi_2$ are necessarily satisfied, but the opposite is not true.
Since $\epsilon_1$ and $\epsilon_3$ share some facts with different degrees (both have a cup in front of a glass, and a cup in front of another), the graph $M_t$ involves two edges with low weight.
In general, if two nodes $\Phi_i$ and $\Phi_j$ are connected by two opposite edges with weight 1, it implies that $\Phi_i\equiv\Phi_j$.

\subsection{Classification} 
\label{sec:classifying}
The classifying function $\mathcal{C}$ is based on a query to the fuzzyDL reasoner, through which we obtain all the concepts having $\epsilon$ as a realisation, \ie all $\Phi_j\sqsubseteq\Phi$ such that $p^{\class{\epsilon}{\Phi_j}}\neq 0$.
Therefore, the classification is a sub-graph of $M_t$, which we call \emph{classification graph} $M^\star$.
Based on \eqref{eq:learning}, the solution to such a query is the conjunction among the $p_{zh}$ degrees obtained by computing the membership of $c_{zh}$ in the restriction ${^{a}\Omega(k_j)}$ of $\Phi_j$; as deducible from Figure~\ref{fig:leftShoulder}.
Thus, the \emph{classification degree} $p^\class{\epsilon}{\Phi_j}$ is the minimum $p_{zh}$ value among all the \nTH{zh} restrictions of $\Phi_j$.
Such a degree is associated with each node of $M^\star$ to represent whether the beliefs of $\epsilon$ satisfy the restrictions of $\Phi_j$.
If the result to the query is only the root node, \ie $M^\star\equiv\Phi$, then $\epsilon$ is considered as not classified.

For instance, Figure~\ref{fig:exGraph} highlights the nodes of $M_t$ that are part of $M^\star$ when we classify $\epsilon_4$ (Figure~\ref{fig:exScene4}) after having learned and structured $\epsilon_1$ (Figure~\ref{fig:sceneEx}), $\epsilon_2$ and $\epsilon_3$ (Figure~\ref{fig:exScene2} and \ref{fig:exScene3}) as $\Phi_1$, $\Phi_2$ and $\Phi_3$, respectively.
In accordance with Figure~\ref{fig:exTable}, $\epsilon_4$ is not classified as $\Phi_3$ since the beliefs do not satisfy all the restrictions.
Instead, $\epsilon_4$ is classified in $\Phi_1$ with a degree $p^{\class{\epsilon_4}{\Phi_1}}=0.8$.
Since $\Phi_1\sqsubseteq\Phi_2$, $\epsilon_4$ is also classified in $\Phi_2$ with a higher degree (\ie $p^{\class{\epsilon_4}{\Phi_2}}=1$) because $\Phi_2$ restricts fewer beliefs, \ie its a parent of $\Phi_1$.

From the classification degree, we cannot assess whether a child node in $M^\star$ better represents $\epsilon$ with respect to its parent.
Thus, we introduced the \emph{similarity value} $d^{\Phi_j}_\epsilon$ as the ratio between the restrictions of $\Phi_j$ and the $c_{zh}$ cardinalities of $\epsilon$.   
More formally, we associate each node of $M^\star$ with 
\begin{equation}
    d^{\Phi_j}_\epsilon \triangleq \frac{d^{\Phi_j}}{d_\epsilon} 
    \;\;\text{where}\;\;
    d^{\Phi_j} = \sum_{\mathbf{r}_{zh}}^{\Phi_j}\,k_j,
    \;\;\text{and}\;\;
    d_\epsilon = \sum_{\mathbf{r}_{zh}}^\epsilon\,c_{zh}.
\end{equation}
In the crisp case, it would hold that $d^{\Phi_j}_\epsilon\in[0,1]$ since ${d_\epsilon{\geqslant}d^{\Phi_j}}$ always, otherwise, $\Phi_j$ would not be a node of $M^\star$ because its restrictions are not be satisfied, and we would not compute $d^{\Phi_j}_\epsilon$.
In the fuzzy domain, the similarity value might be \emph{slightly} bigger than $1$ when $c_{zh}$ is within the range defined by the fuzziness parameter $a$.
However, the more the similarity exceeds $1$, the lower $p^{\class{\epsilon}{\Phi_j}}$ becomes.

As an example, Figure~\ref{fig:exTable} shows that the similarity between $\epsilon_4$ (Figure~\ref{fig:exScene4}) and $\Phi_1$ (Figure~\ref{fig:sceneEx}) is high (\ie $d^{\Phi_1}_{\epsilon_4}=0.9$) since the category restricts most of the cardinality of the scene.
Instead, the similarity with $\Phi_2$ (Figure~\ref{fig:exScene2}) is lower (\ie ${d^{\Phi_2}_{\epsilon_4}=0.4}$) because the category restricts approximately half of the cardinality of the scene.
The similarity with $\Phi_3$ is undefined since it does not belong to $M^\star$, \ie ${p^{\class{\epsilon_4}{\Phi_3}}=0}$.

\section{Memory Capabilities Implementation}
\label{sec:mem_cap}

This section presents our main contribution, which concern the implementation of a cognitive-like memory based on the SIT algorithm.
In particular, the storing and retrieving functionalities of the memory are based on all the functionalities of SIT, while consolidating and forgetting only exploit structuring.
These memory functionalities are also used to manage the scores assigned to each categories in the memory.

\IncMargin{.2em}
\begin{algorithm}[t]
   \caption{The store or retrieve implementation.}
   \label{alg:storeRetrieve}%
\footnotesize
\KwState{~An ontology representing the memory $M_{t-1}$.}
\KwInput{~The input facts $F$ based on $\mathbf{r}$ and $\Gamma$ (Section~\ref{sec:input}).}
\KwOuput{~It updates the ontology for representing $M_t$.}
\KwParam{~The initial score $q^0\in\mathbb{R}^+$. The fuzziness value $a\in[0,1]$. The learning $u,v\in[0,1]$ and the consolidating $e,f\in[0,1]$ thresholds for $p^{\class{\epsilon}{\Phi_j}}$ and $d^{\Phi_j}_\epsilon$ respectively.}
%
\tcp*[h]{Encode and classify the input Scene.}\\ 
$\epsilon \gets \mathcal{E}(F)$\\
$M^\star  \gets \mathcal{C}\big(M_{t-1},\; \epsilon\big)$\label{ln:classify}\\
%
\tcp*[h]{Add a new CATEGORY into the memory.}\\ 
\If(\TCP{Not classified.}){$M^\star\equiv\{\Phi\}$}{%
  $M_t \gets \mathcal{S}\Big(M_{t-1},\;\mathcal{L}\big(\epsilon,\; q^0,\; a\big)\Big)$\label{ln:learn1}
}\ElseIf{$\big(p^{\class{\epsilon}{\Phi_j}}{<}u ~\&~ d^{\Phi_j}_\epsilon{<}o\big)~\forall\langle\Phi_j, q_j, p^{\class{\epsilon}{\Phi_j}}, d^{\Phi_j}_\epsilon\rangle {\in} M^\star$}{\label{ln:checkLearn2}
  \tcp*[h]{Classified with low degree.}\\
  $M_t \gets \mathcal{S}\Big(M_{t-1},\;\mathcal{L}\big(\epsilon,\; q^0{\cdot}\max_{\Phi_j}^{M^\star}\{q_j\},\; a\big)\Big)$\label{ln:learn2}
}
%
\tcp*[h]{Consolidate if the Scene persists.}\\
\ForEach{$\langle\Phi_j,\; q_j,\; p^{\class{\epsilon}{\Phi_j}},\; d^{\Phi_j}_\epsilon\rangle \in M^\star$}{\label{ln:consIt}
  \If{$\big(p^{\class{\epsilon}{\Phi_j}} > e ~~~\&~~~ d^{\Phi_j}_\epsilon > f\big)$}{\label{ln:checkCons}
    $q_j \gets q_j + p^{\class{\epsilon}{\Phi_j}}$\label{ln:cons}\TCP{Increase $M_t$ scores.}
  }
}
\end{algorithm}

\subsection{Storing and Retrieving}
\label{sec:store}
Algorithm~\ref{alg:storeRetrieve} implements the storing and retrieving functionalities, which are based on the same procedure with the differences that
($i$)~input facts $F$ are perceived or requested, respectively, and for retrieving
($ii$)~$M^\star$ is \emph{returned} and
($iii$)~$\mathcal{L}$ might not be used.
Algorithm~\ref{alg:storeRetrieve} uses all SIT functionalities to classify the input facts and, if necessary, to learn a new scene category.
Then, Algorithm~\ref{alg:storeRetrieve} increases the scores of some classified categories since we assume that scenes persistently observed deserve more attention than sporadic ones.
In our experiments (detailed in Section~\ref{sec:case2}), we considered only the storing functionality, but the results related to retrieving are compliant with the capability of SIT~\cite{l2018DialogueBased}.
It is noteworthy that with this implementation we would also consolidate scenes that are often requested to be retrieved.

In particular, Line~\ref{ln:classify} encodes facts to obtain the beliefs of $\epsilon$, which are used for classification.
If SIT cannot classify the scene, Line~\ref{ln:learn1} learns and structures a new category in the memory $M_t$, which then contains a new node with an initial score $q^0$.
At Line~\ref{ln:checkLearn2} the classification is considered not accurate enough if for all the nodes in $M^\star$ the classification degree is less than a threshold $u$, and the similarity value is less than $o$.
In this case, Line~\ref{ln:learn2} adds a new category to $M_t$, and it exploits a heuristic to set its initial score as the maximum score of the classified categories weighted by $q^0$.
Then, Line~\ref{ln:consIt} iterates over each node of $\Phi_j\in M^\star$ to increase the score $q_j$ if the classification is not negligible, \ie the classification degree is greater than a threshold $e$, and the similarity value is greater than $f$ (Line~\ref{ln:checkCons}).
If those conditions are true, Line~\ref{ln:cons} implements a heuristic that increases the score of $\Phi_j$ by adding its classification degree.
Hence, Line~\ref{ln:learn2} learns categories based on previous scores, while Line~\ref{ln:cons} consolidates recurrent scenes and sub-scenes.

\IncMargin{.3em}
\begin{algorithm}[t]
   \caption{The consolidate and forget implementation.}
   \label{alg:consolidateForget}%
\footnotesize
\KwState{~An ontology representing the memory $M_{t-1}$.}
\KwOuput{~It updates the ontology for representing $M_{t}$.}
\KwParam{~The score weight $l\in\mathbb{R}^+$. The threshold for weak score $g\in[0,1]$ to forget. A \emph{forgotten} flag initialised to false.}
%
\tcp*[h]{Weight the scores.}\\
\ForEach{$\langle\Phi_j,\; q_j\rangle \in M_{t-1}$}{\label{ln:itW}
  $q_j \gets l{\cdot}q_j$\label{ln:consW}
}
$q_{\text{max}} \gets \max_{\Phi_j}^{M_{t-1}}\{q_j\}$\\\label{ln:max}
\ForEach{$\langle\Phi_j,\; q_j\rangle \in M_{t-1}$}{
  $q_j \gets q_j{}/q_{\text{max}}$\label{ln:norm}\TCP{Normalize the scores.}\\
  \If(\TCP{Eventually forget.}){$q_j < g$}{\label{ln:checkForg}
    $M_{t-1} \gets M_{t-1}\setminus\langle\Phi_j,\, q_j\rangle$\\\label{ln:forg}
    $\emph{forgotten} \gets \emph{true}$
  } 
}
%
\tcp*[h]{Eventually restructure the memory.}\\
\If{~~\normalfont\emph{forgotten}~~}{
  $M_t \gets \mathcal{S}\big(M_{t-1},\;\emptyset\big)$\label{ln:restr}
}\Else{
  $M_t \gets M_{t-1}$\label{ln:upScore}
}
\end{algorithm}

\begin{figure*}
    \footnotesize
    \centering~%
    \begin{minipage}[c]{0.4\linewidth}%
        \strut\vspace*{-\baselineskip}\newline
        \centering
        \subfloat[$\epsilon_{10}$.]{%
            \centering%
            \includegraphics[width=0.2\textwidth]{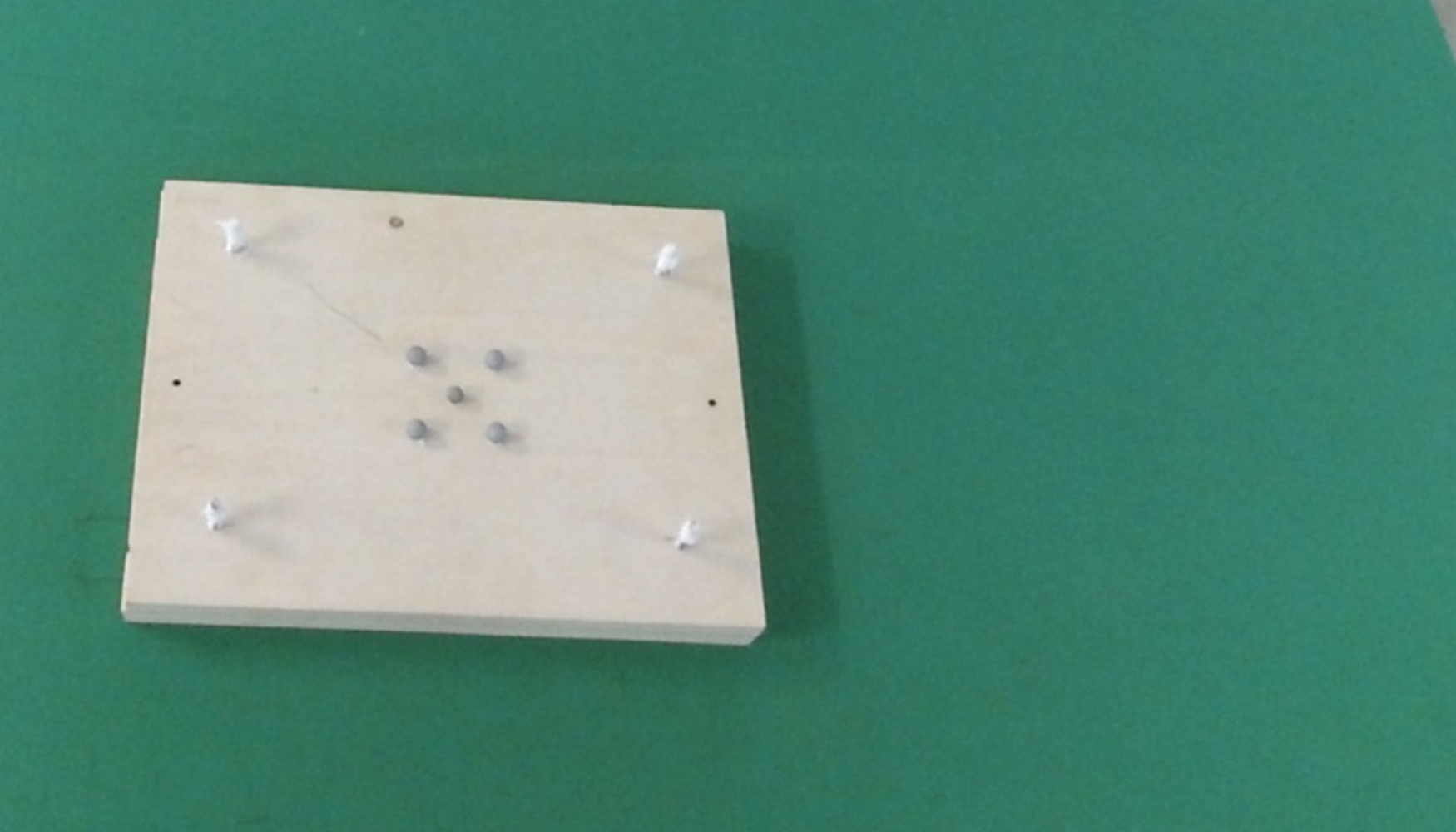}%
            \label{fig:memResult1}%
        }%
        ~%
        \subfloat[$\epsilon_{19}$.]{%
            \centering%
            \includegraphics[width=0.2\textwidth]{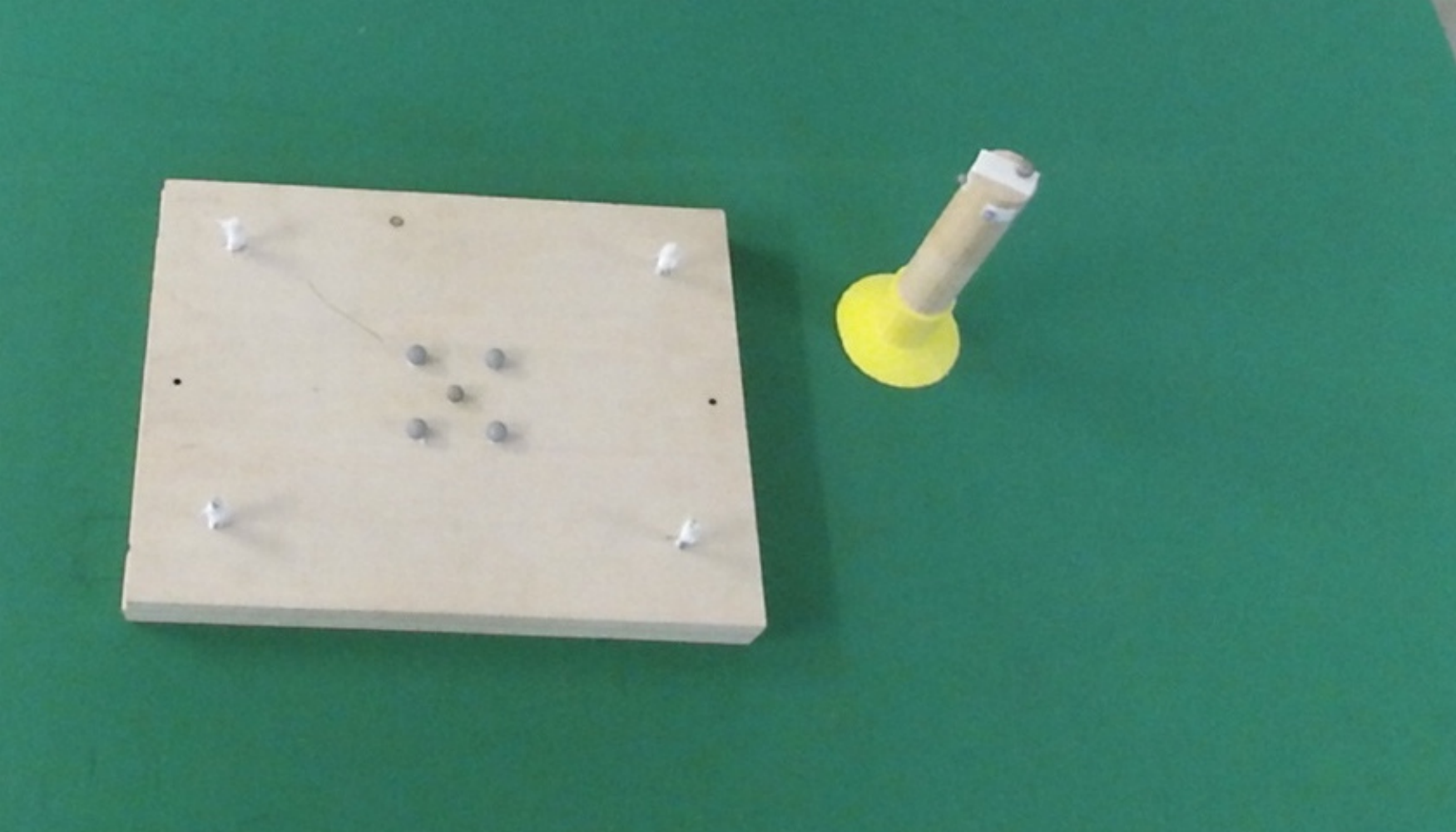}%
            \label{fig:memResult2}%
        }%
        ~%
        \subfloat[$\epsilon_{21}$.]{%
            \centering%
            \includegraphics[width=0.2\textwidth]{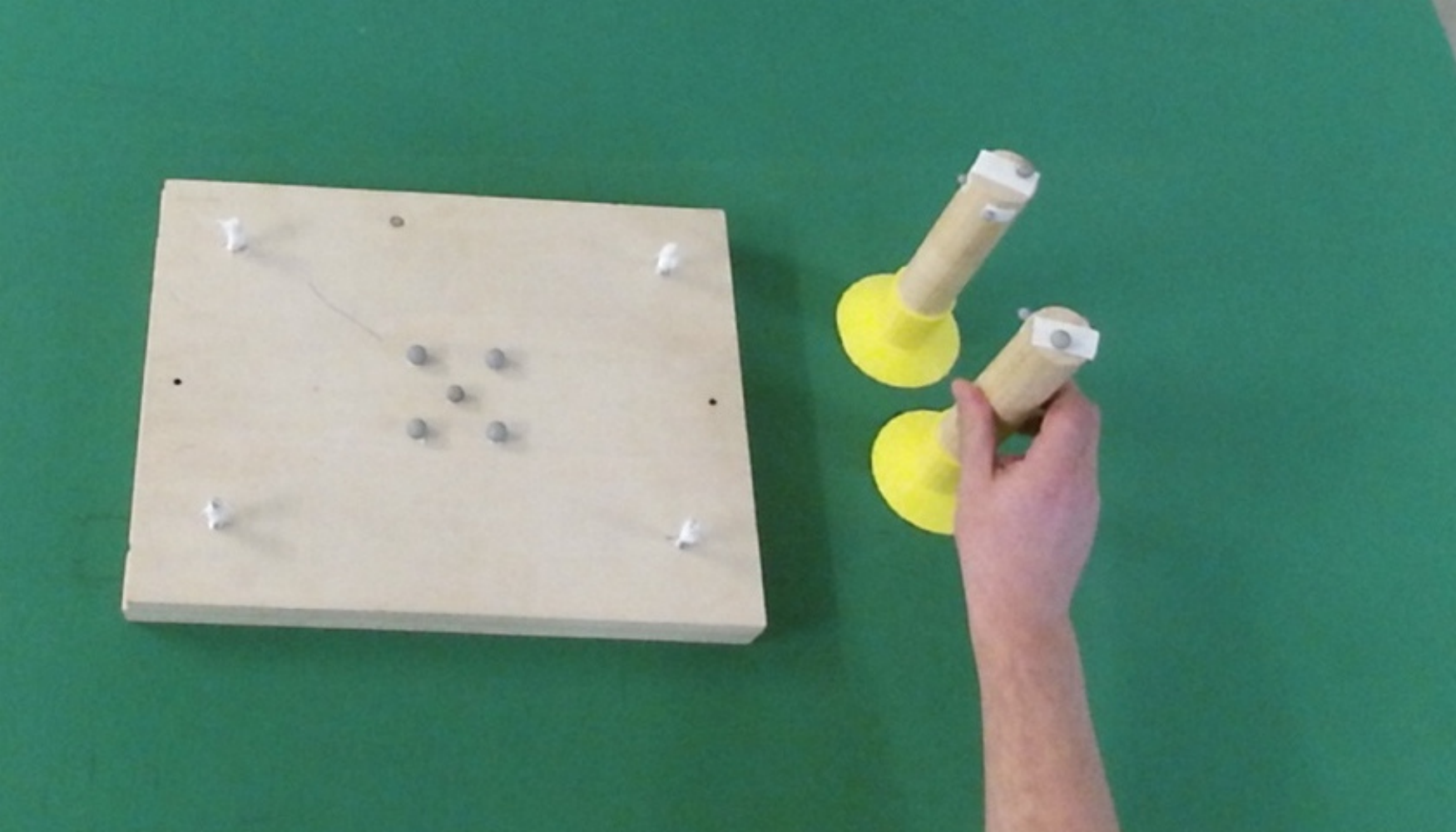}%
            \label{fig:memResult3}%
        }%
        ~%
        \subfloat[$\epsilon_{25}$.]{%
            \centering%
            \includegraphics[width=0.195\textwidth]{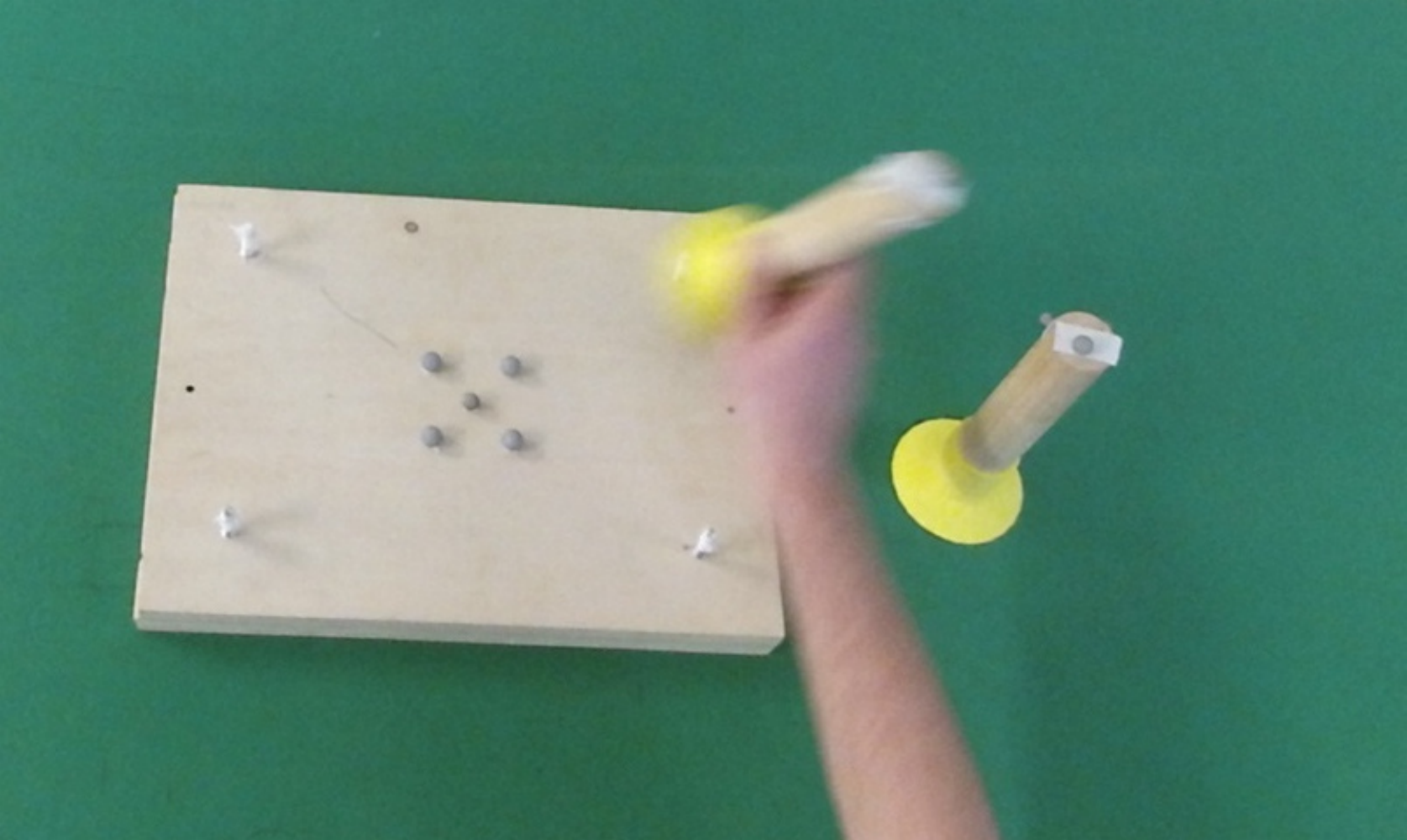}%
            \label{fig:memResult4}%
        }%
        ~%
        \subfloat[$\epsilon_{26}$.]{%
            \centering%
            \includegraphics[width=0.2\textwidth]{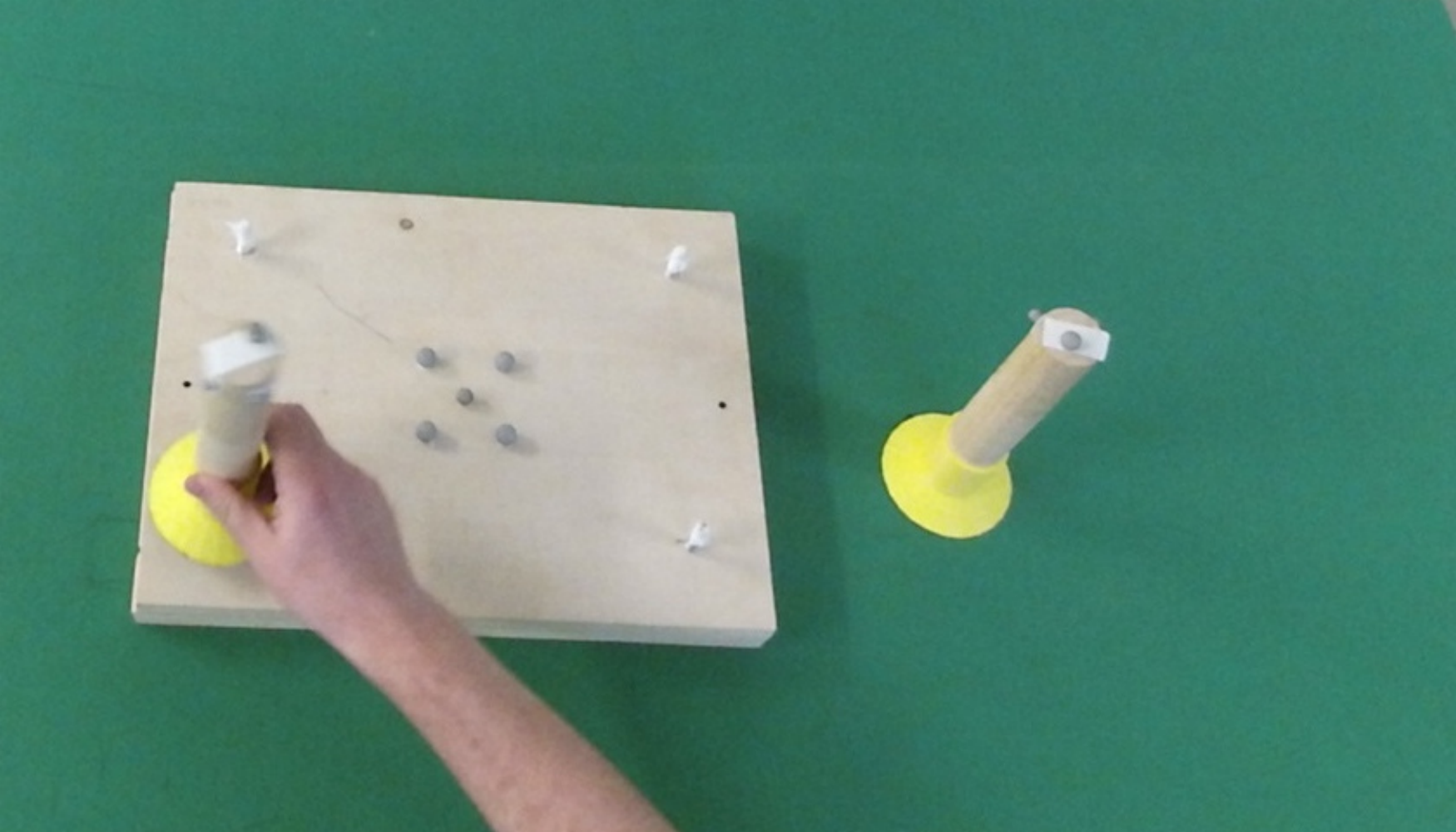}%
            \label{fig:memResult5}%
        }%
        
        \vspace{.4em}
        \subfloat[$\epsilon_{33}$.]{%
            \centering%
            \includegraphics[width=0.2\textwidth]{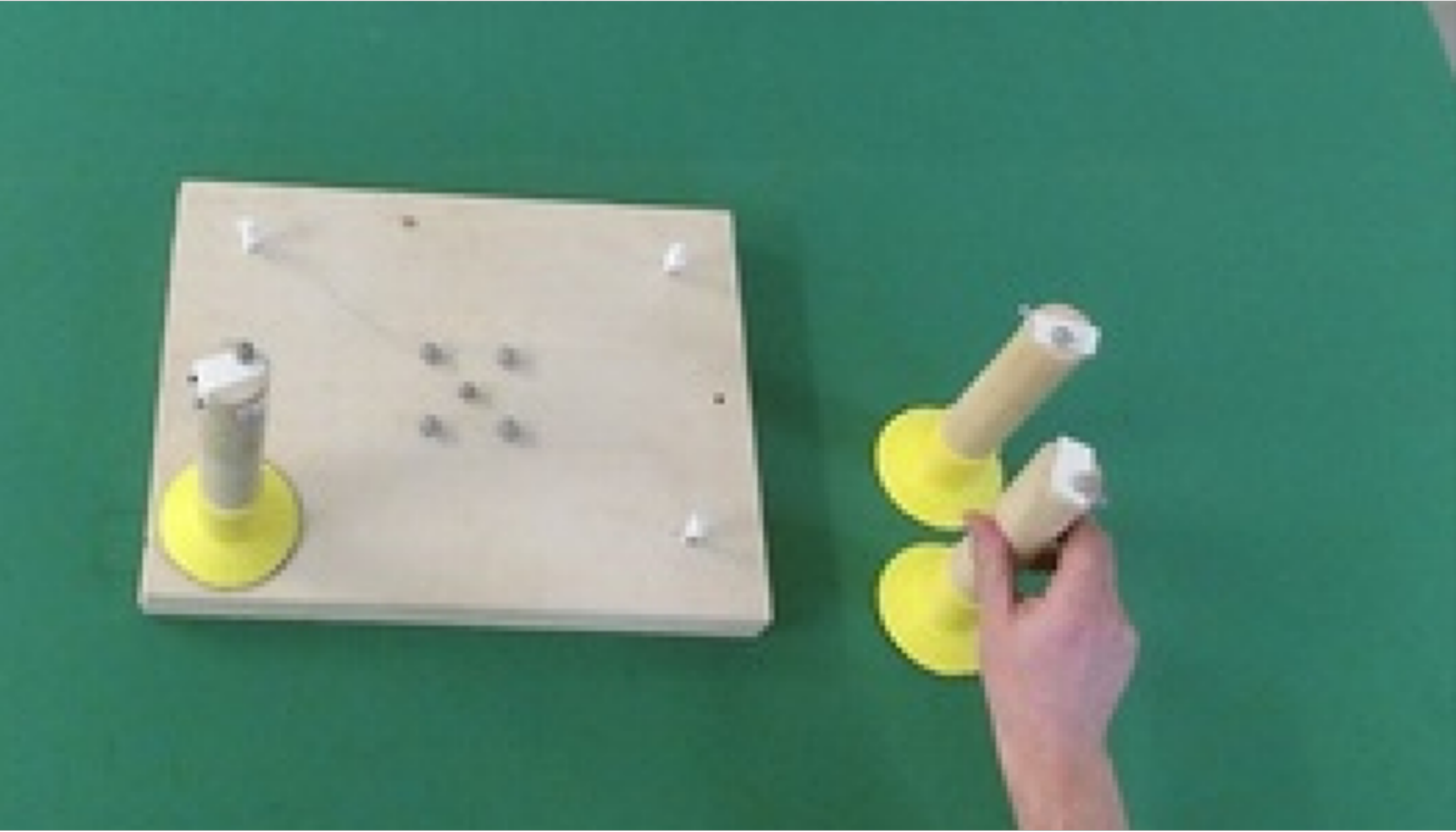}%
            \label{fig:memResultB}%
        }%
        ~%
        \subfloat[$\epsilon_{38}$.]{%
            \centering%
            \includegraphics[width=0.2\textwidth]{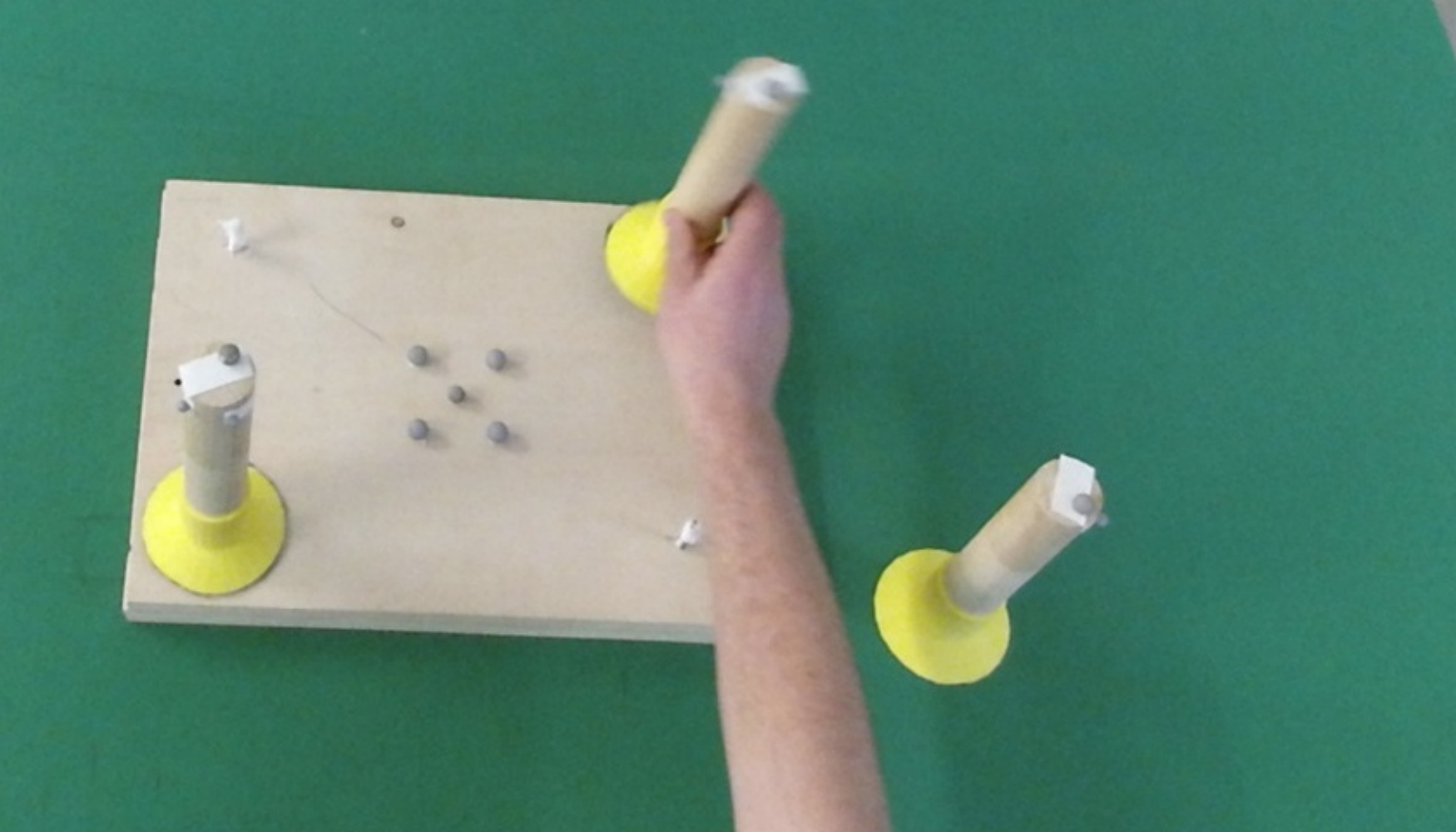}%
            \label{fig:memResult6}%
        }%
        ~%
        \subfloat[$\epsilon_{48}$.]{%
            \centering%
            \includegraphics[width=0.2\textwidth]{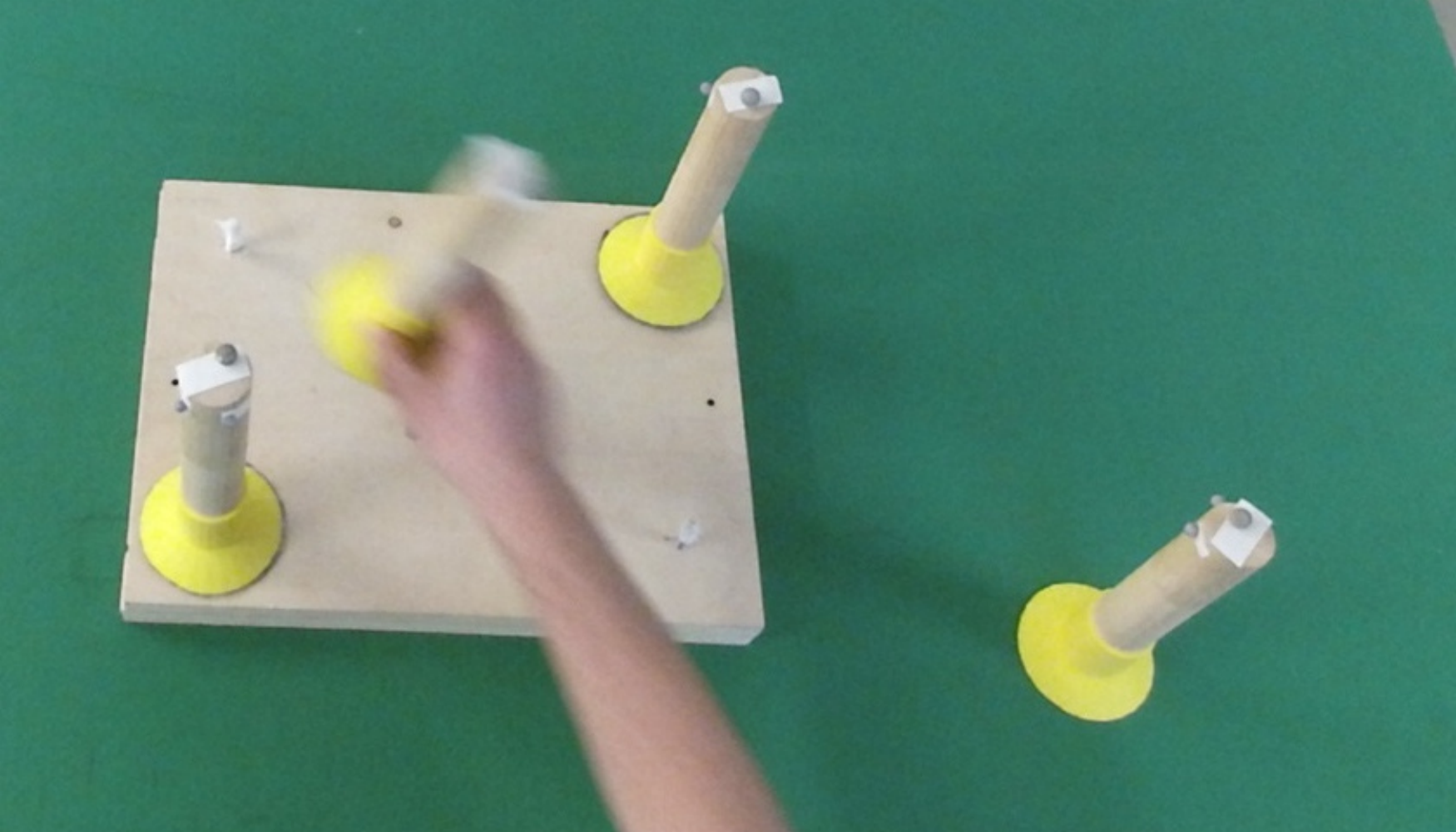}%
            \label{fig:memResult7}%
        }%
        ~%
        \subfloat[$\epsilon_{49}$.]{%
            \centering%
            \includegraphics[width=0.2\textwidth]{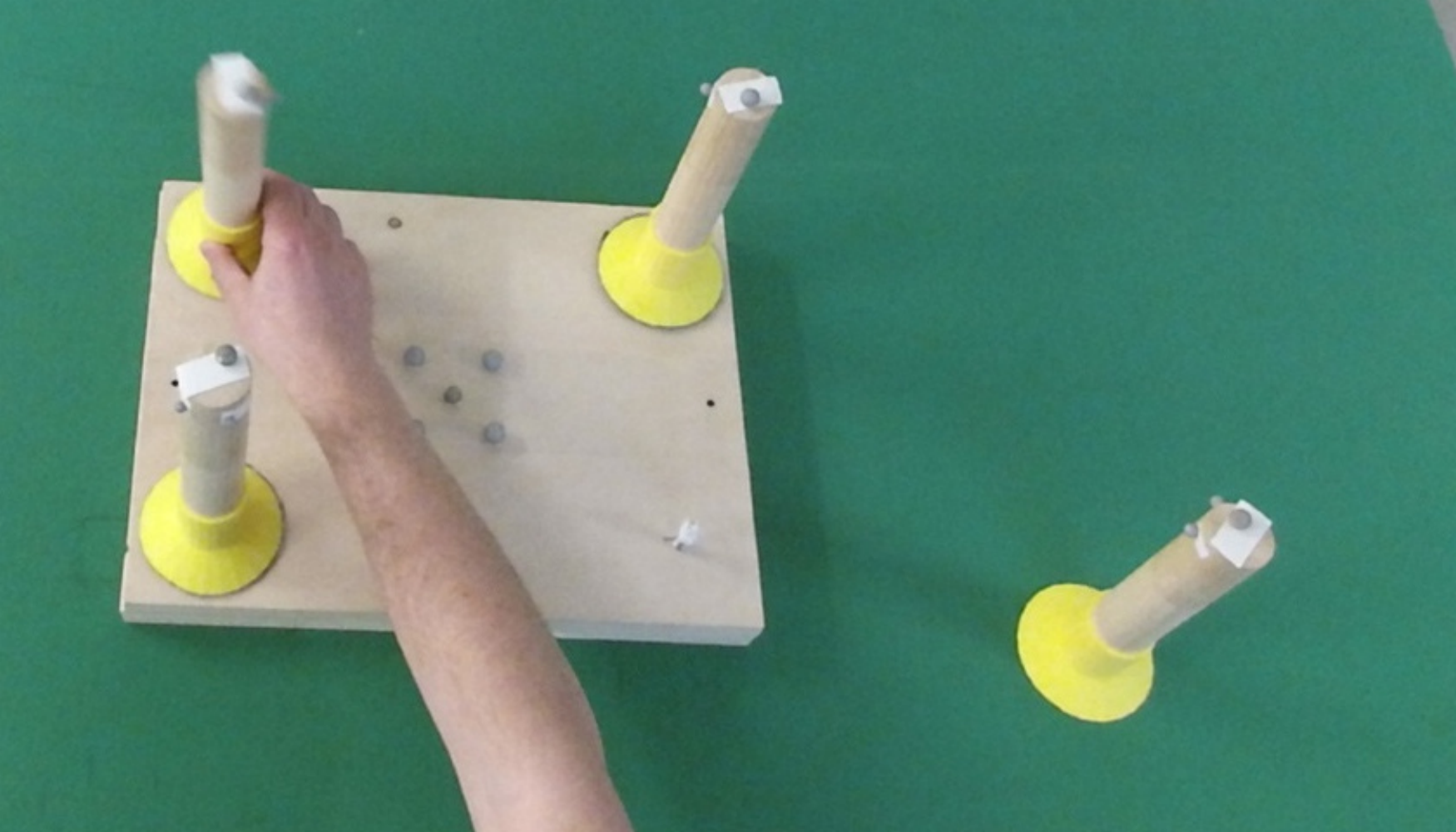}%
            \label{fig:memResult8}%
        }%
        ~%
        \subfloat[$\epsilon_{53}$.]{%
            \centering%
            \includegraphics[width=0.2\textwidth]{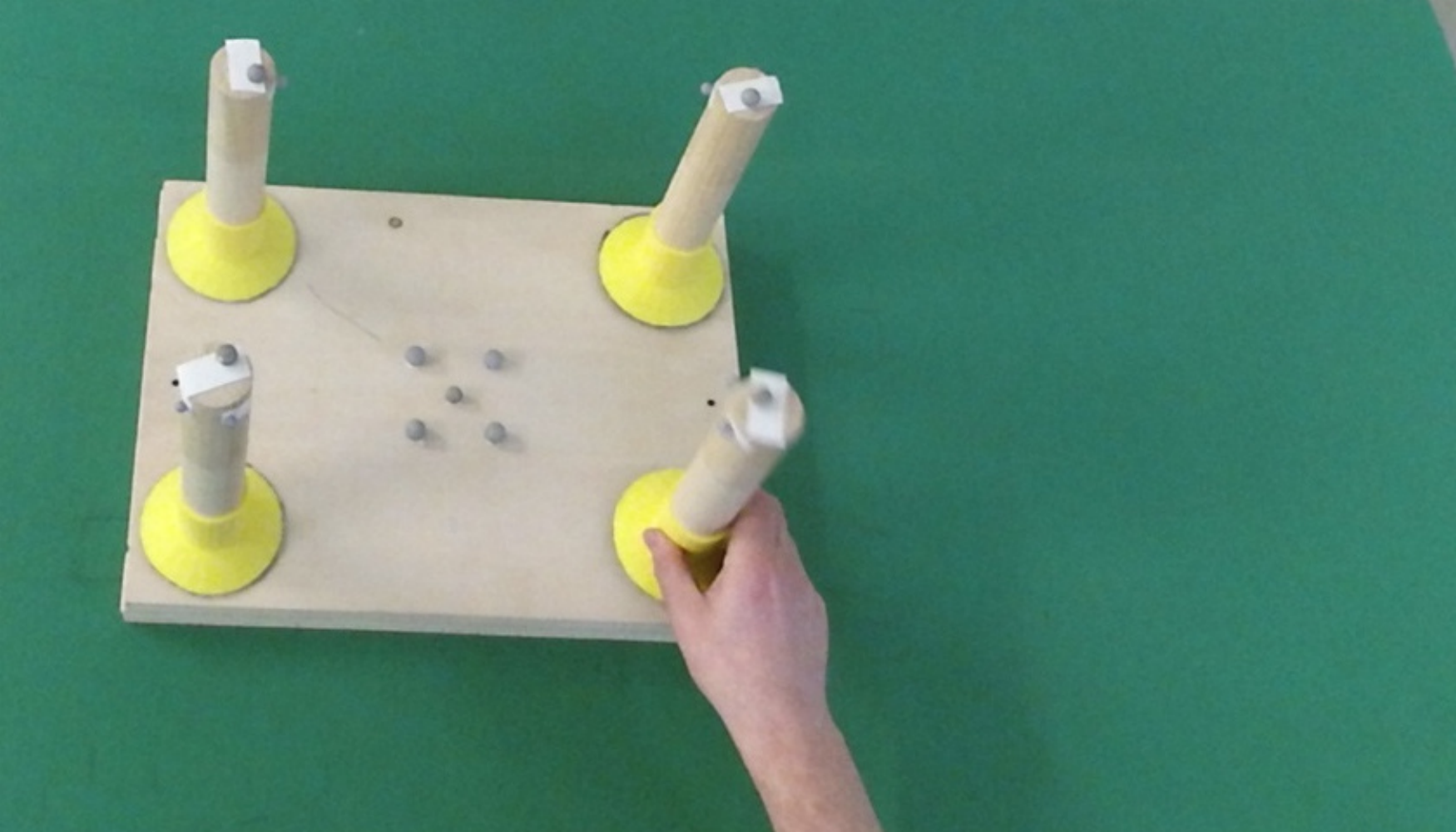}%
            \label{fig:memResult9}%
        }%
    \end{minipage}
    \hfill
    \begin{minipage}[c]{0.58\linewidth}%
        \centering
        \subfloat[Learned scene categories $\Phi_j$ and their consolidating score $q$.]{%
            \rawepstextikz{0.98}{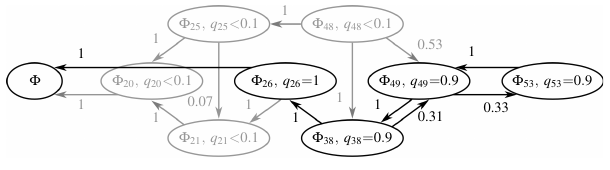}{\centering\scriptsize\strut\vspace*{-\baselineskip}\newline}%
            \label{fig:memResultGraph}%
        }%
    \end{minipage}        
    \caption{%
        The demonstration of a table assembly task and the bootstrapped representation. 
        Figures~(\ref{fig:memResult1}-\ref{fig:memResult9}) show some of the scenes perceived during the demonstration, and Figure~(\ref{fig:memResultGraph}) shows the representation bootstrapped in the robot's memory when the demonstration ended. 
        For clarity, the figures denote scenes $\epsilon$ and categories $\Phi$ with an index related to the time instant of the demonstration.
        Also, each scene category encodes a symbolic representation of the related scene (\ie how many legs have been connected to the table), which is not shown in the figures.
        Note that the grey categories in the memory graph have been forgotten, while the other nodes show the scenes that have been consolidated, \ie the persistent scenes and sub-scenes that are relevant for bootstrapping a representation of the assembly task.
    }%
    \label{fig:tableMount}%
\end{figure*}%

\subsection{Consolidation and Forgetting}
\label{sec:consolidate}
Algorithm~\ref{alg:consolidateForget} shows the implementation of the consolidating and forgetting strategies, and it is sporadically used to update the score $q_j$ of each $\Phi_j$ in $M_{t-1}$.
If a score is too low, Algorithm~\ref{alg:consolidateForget} removes the related scene category from the memory, and it structures a new graph $M_t$.
Otherwise, it generates a graph $M_t$ that has the same structure of $M_{t-1}$, but the nodes would generally have different scores.

At Line~\ref{ln:itW}, we iterate for each node in the memory, and we implement a heuristic to update all the scores.
In particular, Line~\ref{ln:consW} weights each score $q_j$ with a constant value $l$ and, Line~\ref{ln:norm} divides them by the maximum score in $M_t$ to normalise each $q_j$ in $[0,1]$.
Then, if the normalised value is lower than a threshold $g$ (Line~\ref{ln:checkForg}), we consider $\Phi_j$ to have a weak consolidation, and Line~\ref{ln:forg} removes $\Phi_j$ from $M_t$.
Then, if some categories have been forgotten, Line~\ref{ln:restr} invokes the structuring function without specifying the new category to add, \ie it invokes the fuzzyDL reasoner to remove the edges related to forgotten nodes in the graph.
Otherwise, Line~\ref{ln:upScore} only assigns to the memory $M_t$ the graph $M_{t-1}$ with the updated scores.

It is noteworthy that we do not implicitly decrease the scores $q_j$, but this happens as a consequence of the normalisation, which is needed to compare $q_j$ with a forgetting threshold $g$ to be set \emph{a priori}.
However, such a normalisation generates scores that are discontinuous over time, which lead to issues, as discussed in Section~\ref{sec:discussion}.

Also, Algorithm~\ref{alg:consolidateForget} minimizes the usage of $\mathcal{S}$ (by the means of the \emph{forgotten} flag) since it is a time-consuming task that scales exponentially with the number of nodes in $M_{t-1}$.
This also holds for $\mathcal{C}$ because both exploit the fuzzyDL reasoner, which makes Algorithm~\ref{alg:storeRetrieve} more complex than Algorithm~\ref{alg:consolidateForget}.

\section{Memory Evolution During a Demonstration}
\label{sec:case2}

\subsection{Experimental Setup}
SIT was configured with an input interface involving one symmetric relation $\mathbf{r}\sqsupseteq\{\onto{connected}\}$, and two types of element
${\Gamma\sqsupseteq\{\onto{CONNECTOR}, \onto{LEG}\}}$.
From point clouds, we perceived the centre of mass of the objects projected in the workbench.
Given the 2D position of two objects associated with $\gamma_x$ and $\gamma_y$, we considered them connected with a degree that is linearly dependent on their distance $d_{xy}$, \ie \fuzzyprop{\gamma_x}{\gamma_y}{\mathbf{r}_1}{1} if $d_{xy}{=}0$, whereas $p^{\prop{\gamma_x}{\gamma_y}{\mathbf{r}_1}}{=}0$ if $d_{xy}{>}0.15$m.

The robot observed a human performing an assembly task involving a \emph{table} having four connecting points, \ie each one is a \onto{CONNECTOR} where a \onto{LEG} should be fixed.
The objects were placed in unknown locations and order.
The task was completed when all the legs were properly placed.
From a demonstration, we collected $68$ scenes%
    \footnote[4]{We evaluate a dataset available at \url{https://github.com/buoncubi/HRI_tasks_dataset}, where other executions of similar tasks are also concerned.} 
with a rate of $2.5$Hz (some of which are shown in Figure~\ref{fig:tableMount}) that were encoded in $\epsilon_1\ldots\epsilon_{68}$ instances (through Line~\ref{ln:classify}).

With a fuzziness of $a=0.4$, and an initial score of ${q^0=0.5}$, we set Algorithm~\ref{alg:storeRetrieve} to store new categories even if the classification had a high degree (\ie $u=0.9$) and similarity (\ie $o=0.8$).
Hence, we tended to store many categories in the memory, and we relied on consolidating and forgetting functionalities to avoid over-fitting, \ie maintaining only the relevant scenes for the task.
Also, we consolidated scenes that are well classified (\ie $e=0.9$), even if partially (\ie $f=0.2$) to consolidate sub-scenes.
We invoked Algorithm~\ref{alg:consolidateForget} with $l=10$ and $g=0.1$ to evaluate scores and forget weak categories every five encoded scenes.

\subsection{Store, Consolidate, and Forget Memory Items} 
Figure~\ref{fig:memResultGraph} shows the task, as well as the consolidated and forgotten categories, as stored in the memory of the robot after the demonstration.
In particular, it is possible to derive an undirected graph having only the edges with a weight equal to $1$, and we can deduce the assembly task as the sequence $\Phi_{26}$, $\Phi_{38}$, $\Phi_{49}$ and $\Phi_{53}$, where in
($i$)~$\Phi_{26}$ a leg is connected, \emph{i}.\emph{e}., \defff{\reify{connected}{LEG}}{^a\Omega(.99)},
($ii$)~$\Phi_{38}$ \emph{almost two} legs are connected, \emph{i}.\emph{e}., \defff{\reify{connected}{LEG}}{^a\Omega(1.85)},
($iii$)~$\Phi_{49}$ \emph{three} legs are connected, \emph{i}.\emph{e}., \defff{\reify{connected}{LEG}}{^a\Omega(2.55)},
($iv$)~$\Phi_{53}$ \emph{four} legs are connected, \emph{i}.\emph{e}., \defff{\reify{connected}{LEG}}{^a\Omega(3.47)}.
The restriction was lower than $3$ and $4$ due to perception vagueness, \ie on average $p^{\prop{\gamma_x}{\gamma_y}{\mathbf{r}_1}}=0.75$ instead of~$1$.

Until the \nTH{20} scene, the distances among objects was too high, and SIT was not provided with any inputs, \ie $p^{\prop{\gamma_x}{\gamma_y}{\mathbf{r}_1}}=0$. 
The \nTH{20} scene represented a configuration having two legs connect between each other with a restriction of $\Omega(0.09)$, which increases to $\Omega(0.73)$ in $\Phi_{21}$ since the human moved the second leg closer to the first, as shown in Figure~\ref{fig:memResult7}.
Similarly, we stored categories $\Phi_{25}$ (Figure~\ref{fig:memResult4}) and $\Phi_{48}$ (Figure~\ref{fig:memResult7}).
However, those categories were forgotten since the related scenes were not persistent during the demonstration, \ie~${\{\Phi_{20},\Phi_{21},\Phi_{25},\Phi_{48}\}}$ did not occur in $M^\star$ often, and Line~\ref{ln:learn2} would not be used to consolidate their score.
Instead, SIT persistently classified $\{\Phi_{26},\Phi_{38},\Phi_{49},\Phi_{53}\}$ and, through Line~\ref{ln:cons}, we increased their score. 
Due to normalisation (Line~\ref{ln:norm}), the not consolidated categories get a lower score with respect to the consolidated ones, until they become weak, and then forgotten through Lines~\ref{ln:forg} and \ref{ln:restr}.

\subsection{Computation Performance} 
According to Section~\ref{sec:classifying}, we observed a similarity value $d^{\Phi_j}_\epsilon>1$ depending on the fuzziness parameter $a$, the state of the memory $M_t$, and the current scene $\epsilon$.
We observed this event $37$ times over $134$ measurements with $d^{\Phi_j}_\epsilon < 1.036$.

The computation required by an Intel i5 ($2.53$GHz) to compute Algorithm~\ref{alg:storeRetrieve} was $0.55\pm 4.5$s on average.
The variance was due to OWL reasoning complexity demanded by the structuring and classification functions.
The worst-case reasoning time strongly limited SIT, but the average computation time was significantly lower.
Noteworthy, the characterisation of incremental OWL reasoning time is far from trivial also because it dependents on previous states.
Therefore, to properly characterise the average complexity, a systematic evaluation of different memory graphs is required.

We did not observe any inconsistency when the scenes were perceived up to $0.31$Hz (\ie 8 times slower) since we were able to sample and process all the relevant scenes of the demonstration, which held for 27.2s.
Thus, the ability to compute our framework online is also related to synchronisation issues concerning human speed during the demonstration.

\section{Discussion}
\label{sec:discussion}

We previously pointed out that our framework can address all the requirements presented in Section~\ref{sec:relatedWork}.
Indeed, it can learn structured models (\ie it bootstraps graphs of categories in the ontology) from non-annotated data, which are generated through a single demonstration.
Since we used a symbolic formalism, the learned models are intelligible, and they can be refined during the demonstration.
However, these models should be presented in a simplified form to result more intuitive to a human. 
Also, we showed that our framework can bootstrap knowledge while the robot observes the demonstration, \ie it can be used online.
However, our framework is limited by the OWL reasoning complexity since it increases exponentially with the size of the memory graph and the number of symbols in the input interface, \ie $w$ and $v$.

Thanks to the OWL symbolic formalisation, we could base the framework on a general-purpose input interface, which should be carefully designed \emph{a priori}.
In order to support a general-purpose design, we have shown how the learned model changes based on two different contexts addressed in Figures~\ref{fig:SITex} and \ref{fig:tableMount}.
The main limitation of the input interface is that it requires elements of the environment that are related to each other.
Nonetheless, the symbolic formalisation supports the extension to more complex semantics (\eg reify colour and affordance: \reify{in}{\reify{BLUE}{CONTAINER}}) with the drawback of increasing the computation complexity.
The framework can be used with specific input interfaces to model different contexts from the same observations.
This allows implementing different types of memory, such as short-term or long-term memory, which also involve declarative memories, \eg semantic and episodic~\cite{SQUIRE2004171}.

To use our framework in a real application, we need other components to map the numerical data of sensors and actuators into symbols.
With a fuzzy formalisation of symbols, we have shown that the framework is robust to vague perception, but a drawback is that a cardinality restriction based on the $\sigma$-count approach might be ambiguous.
Since we based our system on OWL, we can rely on a flexible formalism and exploit other reasoning mechanisms as well; but this would further limit scalability due to computation complexity.

In Section~\ref{sec:mem_cap}, we implemented a rehearsal-like heuristic approach to consolidate persistent scenes and forget sporadic configurations through a scoring strategy.
We have shown that this approach avoids over-fitting an assembly task demonstration.
However, our heuristics need to be further validated and integrated with other strategies to identify the knowledge to be forgotten, \eg based on the trace decay theory. 
For this purpose, Algorithms~\ref{alg:storeRetrieve} and Algorithm~\ref{alg:consolidateForget} can be extended based on the SIT functionalities, and a systematic evaluation of different heuristics should be undertaken.
In this paper, we used a simple consolidation function based on hand-tuned thresholds, which is hard to generalise.
However, we are confident that a deeper evaluation of different demonstrations for specific scenarios can lead to the design of a more general consolidating strategy.

Last but not the least, our heuristics need to normalise the consolidating scores for each scene since it relies on thresholds that have to be configured for some range.
This is a drawback because it leads to discontinuous scores over time.
Depending on the memory state, the robot might forget a scene when it has just been stored, or it might not forget it if other scenes are not frequently observed.
To overcome this issue, it is possible to rely on a scoring strategy that is bounded and does not require normalisation (\eg that relies on Sigmoid-based heuristics), but this would also require to decrease the scores during the storing and retrieving functionalities explicitly.

\section{Conclusions}
We present a general-purpose framework inspired by cognitive-like memory that encodes, stores, retrieves, consolidates, and forgets scenes on the basis of a learning algorithm that exploits fuzzy symbols into an OWL ontology.
The framework classifies scenes from previous observations, and it bootstraps an intelligible structure of qualitative sub-scenes.
Also, it can online bootstrap knowledge from non-annotated data acquired during a single human-led demonstration.

With a heuristic-based score mechanism, we could learn a symbolic planning domain for an assembly task, and we showed that consolidation and forgetting could be a strategy to avoid over-fitting the demonstration.
We also discussed that our framework can be used to systematically investigate different heuristics (\eg Sigmoid-based) and, as further work, we want to compare them for demonstrations that involve more than one task. 

\bibliographystyle{IEEEtran}
\bibliography{IEEEabrv,bib}

\end{document}